\documentclass[sigconf]{acmart}

\usepackage{amsmath,amsfonts}
\usepackage{algorithmic}
\usepackage{algorithm}
\usepackage{array}
\usepackage[caption=false,font=normalsize,labelfont=sf,textfont=sf]{subfig}
\usepackage{textcomp}
\usepackage{url}
\usepackage{verbatim}
\usepackage{graphicx}
\usepackage{epstopdf}
\usepackage{color, colortbl}
\usepackage{bm}
\usepackage{stfloats}
\usepackage{float}
\usepackage{caption}
\usepackage{multirow}
\usepackage{utfsym}

\usepackage{subfig} 
\usepackage{subfloat}

\usepackage{booktabs}
\makeatletter
\renewcommand{\maketag@@@}[1]{\hbox{\m@th\normalsize\normalfont#1}}%
\makeatother

\begin{document}
\settopmatter{printacmref=false} 
\renewcommand\footnotetextcopyrightpermission[1]{} 
\title{Learning a Graph Neural Network with Cross Modality Interaction for Image Fusion}

\author{Jiawei Li}
\affiliation{%
	\institution{University of Science and Technology Beijing}
	\city{Beijing}
	\country{China}
	}
\email{ljw19970218@163.com}

\author{Jiansheng Chen}
\affiliation{%
	\institution{University of Science and Technology Beijing}
	\city{Beijing}
	\country{China}
}
\email{jschen@ustb.edu.cn}
\authornote{Corresponding author: Jiansheng Chen.}

\author{Jinyuan Liu}
\affiliation{%
	\institution{Dalian University of Technology}
	\city{Dalian}
	\country{China}
}
\email{atlantis918@hotmail.com}

\author{Huimin Ma}
\affiliation{%
	\institution{University of Science and Technology Beijing}
	\city{Beijing}
	\country{China}
}
\email{mhmpub@ustb.edu.cn}

\renewcommand{\shortauthors}{Jiawei Li, Jiansheng Chen, Jinyuan Liu, \& Huimin Ma}


\begin{teaserfigure}
	\centering
	\includegraphics[width=0.98\textwidth]{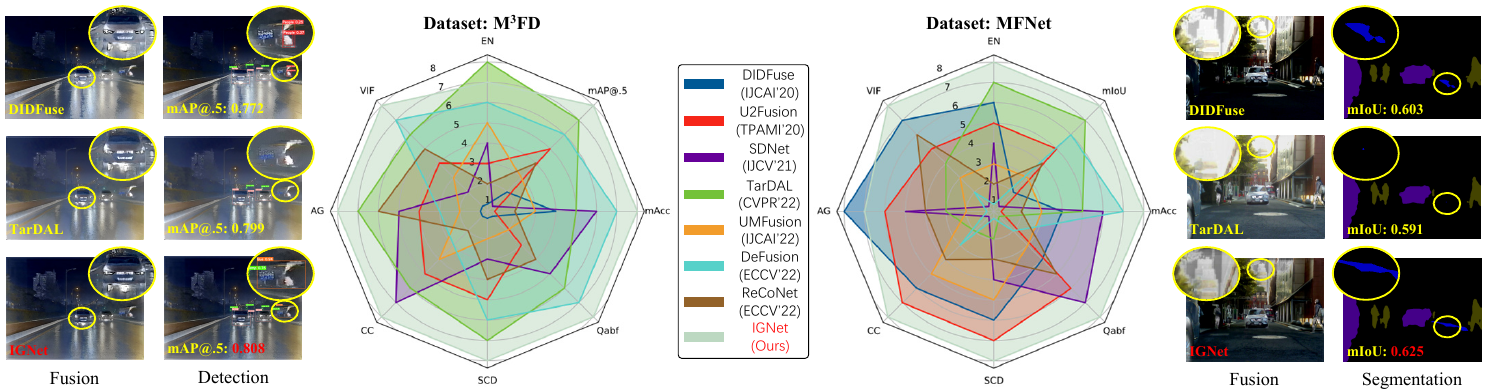}
	\caption{Fusion, detection and segmentation comparisons with state-of-the-art methods on $ \bf{M^3}$FD and MFNet datasets. We can obviously notice the superiority of our IGNet in the zoomed-in patches and radar plots.}
	\label{introduction}	
\end{teaserfigure}

\begin{abstract}
Infrared and visible image fusion has gradually proved to be a vital fork in the field of multi-modality imaging technologies. In recent developments, researchers not only focus on the quality of fused images but also evaluate their performance in downstream tasks. Nevertheless, the majority of methods seldom put their eyes on mutual learning from different modalities, resulting in fused images lacking significant details and textures. To overcome this issue, we propose an interactive graph neural network (GNN)-based architecture between cross modality for fusion, called IGNet. Specifically, we first apply a multi-scale extractor to achieve shallow features, which are employed as the necessary input to build graph structures. Then, the graph interaction module can construct the extracted intermediate features of the infrared/visible branch into graph structures. Meanwhile, the graph structures of two branches interact for cross-modality and semantic learning, so that fused images can maintain the important feature expressions and enhance the performance of downstream tasks. Besides, the proposed leader nodes can improve information propagation in the same modality. Finally, we merge all graph features to get the fusion result. Extensive experiments on different datasets ($ i.e. $, TNO, MFNet, and $\rm M^3$FD) demonstrate that our IGNet can generate visually appealing fused images while scoring averagely 2.59\% mAP@.5 and 7.77\% mIoU higher in detection and segmentation than the compared state-of-the-art methods. The source code of the proposed IGNet can be available at \href{URL} {https://github.com/lok-18/IGNet}.

\end{abstract}



\maketitle

\section{Introduction}

Due to the inadequacy of single-modality imaging, the resulting images are commonly defective in complex scenes \cite{liu2022learn}, \cite{liu2021smoa}. As a representative, Visible images are more in line with the human visual system (HVS), but susceptible to environmental factors. In this case, researchers attempt to fuse visible images with ones of another modality to counteract the disadvantages of single-modality imaging. Complementarily, infrared images can capture salient targets with thermal radiation sensors. Texture details and resolution of them often perform undesirably. Therefore, infrared and visible image fusion (IVIF) emerges as the times require, which can possess information from different modalities simultaneously. Acting as an indispensable part of multi-modality imaging technology, IVIF has drawn extensive attention to computer vision tasks, $ e.g. $, vehicle detection \cite{sun2022drone}, video surveillance \cite{paramanandham2018infrared} and image stitching \cite{jiang2022towards}.

For the past decade, deep learning networks have been introduced to explore the IVIF task \cite{zhang2021image}, which mainly contains convolution neural network (CNN)-based \cite{liu2021learning} and transformer-based methods \cite{ma2022swinfusion}. These methods focus on accurate feature extraction for inputs while promoting the fusion efficiency significantly.
Compared with previous traditional approaches, deep learning-based methods can utilize more efficient feature extraction capabilities to obtain fusion results with higher efficiency. With further development, researchers also pay attention to the performance of down-stream tasks after fusion \cite{tang2023divfusion}. That is to say the results of down-stream tasks are closely related to the fusion images. 


Existing mainstream IVIF methods have reached a certain height, nevertheless, there are still several drawbacks: (\romannumeral1) the uneven distribution of infrared and visible information extracted from networks causes fusion results only to be biased towards one modality \cite{li2023infrared}, which can not perform the prominent regions of source images well. (\romannumeral2) since feature learning often acts separately on each single branch, the information contained in networks may lack the communication of cross modalities \cite{tang2022image}. (\romannumeral3) the internal design of message delivery is not well taken into account in several networks \cite{zhang2023transformer}, so some significant details of source images can not be displayed in fused results.

To alleviate the drawbacks mentioned above, in this paper, we propose an interactive GNN-based architecture between cross modality for the IVIF task, termed as IGNet. Concretely, multi-scale shallow features are first extracted by convolutions and the proposed structure salience module (SSM). Then, we construct a graph interaction module (GIM) to obtain graph structures of different branches for feature learning. Note that the interaction of cross-modality graph features enables the proposed IGNet to achieve more semantic information, which can improve the performance of down-stream tasks, $ e.g. $, object detection, and image segmentation. In addition, the establishment of leader nodes guides the message propagation effectually to avoid image quality degradation caused by feature loss. Fig.~\ref{introduction} proves that our proposed IGNet maintains the superior position regardless of subjective visual results or objective marks compared with state-of-the-art methods. 

In brief, the contributions can be divided into the following aspects:
\begin{itemize}
	\item For optimizing the internal relationship of fusion and down-stream ($ i.e. $, object detection and image segmentation) tasks, to the best of our knowledge, we are the first to apply GNN into the IVIF method. To this end, the fused results can contain faithful visual representation and feature comprehension abilities.
	\item We propose a graph interaction module (GIM) for getting graph structures. It can proceed cross-modality communication through graph features, which highlight the desired details of fusion results. Furthermore, the semantic-wise information can also be extracted by GIM for improving down-stream results.
	\item Unlike the common GNN, the leader nodes are employed for information delivery after achieving graphs. Accompanied by a leader node as a pioneer, fusion images can maintain abundant textures from source inputs.
	\item We conduct image fusion, detection, and segmentation experiments on TNO, $ \rm{M^3}$FD, and MFNet datasets. Compared with the other seven state-of-the-art approaches, our proposed IGNet performs foremost in all tasks.
\end{itemize}

\section{Related Works}
\subsection{Infrared and Visible Image Fusion}
Deep learning has promoted rapid development in the field of image fusion \cite{li2022learning}, \cite{liu2023holoco}, \cite{lei2023galfusion}, \cite{li2023gesenet}. In early stages, researchers are dedicated to improving the performance of fused images by CNN-based methods, which are mainly divided into three classes, $ i.e. $, End-to-End-based models \cite{liu2021learning}, \cite{liu2020bilevel}, Encoder-Decoder models \cite{li2018densefuse}, and generative adversarial network (GAN)-based models \cite{ma2020ddcgan}. 

More specifically, End-to-End models preset parameters before unsupervised training \cite{li2023infrared}. Liu $ et $ $ al. $ \cite{liu2021learning} proposed a coarse-to-fine deep network with an end-to-end manner to learn multi-scale features from infrared and visible images. The structure details were also refined by the proposed edge-guided attention mechanism. The Encoder-Decoder models need to design a fusion rule to integrate features extracted from the encoder, and then output the fusion results from the decoder \cite{xu2022cufd}. Zhao $ et $ $ al. $ \cite{zhao2020didfuse} conducted a novel encoder to decompose source images into background and detail feature maps, which can highlight targets, especially in the dark. The GAN-based models require a generator and a discriminator for adversarial learning. Li $ et $ $ al. $ \cite{li2020attentionfgan} effectively combined the attention mechanism with GAN, namely AttentionFGAN. Moreover, extensive transformer-based models have also received much attention in the IVIF task \cite{ma2022swinfusion}. Tang $ et $ $ al. $ \cite{ma2022swinfusion} utilized Swin Transformer and cross-domain long-range learning into the IVIF task, which connected local features with global representation.

To further explore the performance of fusion images, researchers have introduced down-stream tasks $ e.g. $, object detection and image segmentation, into the IVIF task. As a representative, Liu $ et $ $ al. $ \cite{liu2022target} proposed a unified architecture and built a multi-modality dataset for image fusion and detection. Sun $ et $ $ al. $ \cite{sun2022detfusion} employed the information back-propagated by detection loss in the proposed network to obtain fused images with excellent detection results. For getting more semantic features, Tang $ et $ $ al. $ \cite{tang2022image} proposed a cascaded structure called SeAFusion, which connects the fusion network with a pretrained segmentation module. Zhao $ et $ $ al. $ \cite{zhao2022cddfuse} conducted a novel two-stage training mode for fusion. The detection and segmentation results also performed well in this benchmark.

\subsection{Graph Neural Network}
In recent years, GNN-based approaches have become increasingly popular in computer vision. Different from traditional CNN-based methods, the unique structure of GNN enables to extract and transfer more efficient features. Therefore, GNN is commonly implemented in the feature-wise tasks. As a representative, Xie $ et $ $ al. $ \cite{xie2021scale} proposed a scale-aware network with GNN to conduct few-shot semantic segmentation. In the medical field, Huang $ et $ $ al. $ \cite{huang2021graph} employed a semi-supervised network for medical image segmentation, which could help doctors diagnose diseases better. Recently, GNN becomes popular in the field of saliency detection. It can effectively highlight the salient mask of measured targets. Specifically, Luo $ et $ $ al. $ \cite{luo2020cascade} tried to cascade graph structures for salient object detection (SOD) with RGB-D images. Song $ et $ $ al. $ \cite{song2022multiple} devised a multiple graph module to realize the RGB-T SOD task. GNN can be also applied in Co-Saliency Detection (CSD) and Instance Co-Segmentation (ICS). Li $ et $ $ al. $ \cite{li2021image} presented a general adaptive GNN-based module to deal with CSD and ICS. In addition, some low-level tasks can also perform well by using GNN as their benchmark. Li $ et $ $ al. $ \cite{li2021cross} proposed a novel GNN-based method for image denoising. In summary, GNN maintains sensitivity to semantic information, while handling pixel-level tasks well. Hence, our proposed IGNet can exploit the advantages of GNNs for deeper exploration of IVIF tasks, which can simultaneously improve the quality of fused images and the performance of corresponding downstream tasks.

\begin{figure*}
	\centering
	\includegraphics[width=1\textwidth]{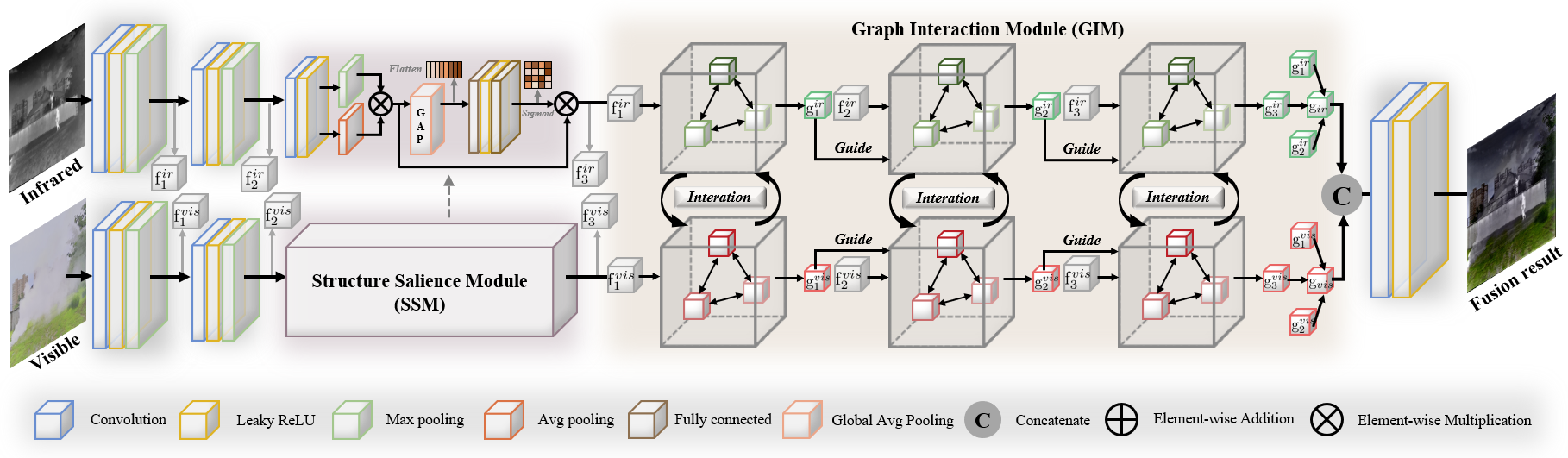}
	\caption{Pipeline of the proposed IGNet. Specifically, we feed multi-scale features into the graph interaction module (GIM) for generating graph structures in different modalities. The cross-modality interaction between graphs is depicted in detail. The leader nodes guide the information delivery from one graph to the latter. Note that we construct graphs in the infrared/visible branch with three loops, respectively. The bottom row represents the legend of the component.}
	\label{network}	
\end{figure*}

\section{Method}
\subsection{Motivation}
In the IVIF task, networks often extract features in infrared and visible branches separately, while ignoring the interaction between modalities. It may cause textures of source images can not be completely displayed in fusion results. With the information delivery during training, the occurrence of feature forgetting is inevitable as well. Besides, the fused images will directly affect the performance of the down-stream results. There is no doubt that applying an effective architecture to achieve visually appealing images can improve the accuracy of detection and segmentation. Significantly, how to obtain fused images with prominent targets, fine textures, and rich semantic information is the key to handling the above issues. Hence, it is our motivation to realize a general IVIF framework, which can obtain fusion and semantic information in pixel and feature domains concurrently.


\subsection{Overall Workflow}

The proposed IGNet adopts a dual-branch framework in the feature learning stages. Subsequently, we aggregate the infrared and visible branches to achieve fusion images. The overall pipeline is illustrated in Fig.~\ref{network}. To be specific, two different-scale features ($ i.e. $, $ {\rm f}^{*}_1 $ and $ {\rm f}^{*}_2 $) can be generated by the first two convolutional layers, where $ * $ denotes the infrared/visible branch. Then, we modify $ {\rm f}^{*}_2 $ through the SSM for getting the salient-structure feature $ {\rm f}^{*}_3 $. It is formulated as follow:
\begin{equation}
	{\rm f}^{*}_3 = \mathcal{S}({\rm f}^{*}_2),
\end{equation}
where $ \mathcal{S} $ means the SSM. For constructing connections between source images, ${\rm f}^{*}_i$ is fed into the GIM to build a learnable graph structure with three loops. We define this process as follows:
\begin{equation}
	{\rm g}_{*} = \sum_{i=1}^{3}{\mathcal{G}({\rm f}^{*}_i)},
\end{equation}
where $ i \in \{1, 2, 3\}$, $ {\rm g}_{*} $ denotes graph features and $ \mathcal{G} $ is the GIM. At last, we combine decorated features to achieve final fusion results:
\begin{equation}
	{\rm I}_f = Conv\big(Concat({\rm g}_{ir}, {\rm g}_{vis})\big),
\end{equation} 
where $ {\rm I}_f $ means fused images. $ Conv(\cdot) $ and $ Concat(\cdot) $ represent convolution and concatenate operations, respectively. Moreover, the employed loss function can effectively transfer information through back-propagation, which is also explicated in Section.~\ref{Loss Function}.

\subsection{Structure Salience Module}

As shown in Fig.~\ref{network}, we use the SSM to optimize $ {\rm f}^{*}_2 $, deepening the expression of deep structure features. After passing through a convolutional layer, the SSM conducts Maxpooling and Avgpooling to coordinate detailed patches and global information simultaneously. We use Element-wise Multiplication to deal with the two pooling information, which can excavate more salient contents from infrared images. Since more detailed textures are contained in the visible branch, Element-wise Addition is exploited to enrich the overall perception instead.

Inspired by SENet \cite{hu2018squeeze}, we also introduce attention to the SSM. Firstly, the aforementioned feature is flattened by Global Average Pooling (GAP). Secondly, we assign two Fully Connected Layers and Sigmoid to generate the corresponding channel weight. It can not only increase the salience of feature representation but also highlight parts that conform to HVS in fused images. Finally, we multiply the feature with channel weight to achieve salient-structure feature $ {\rm f}^{*}_3 $.

\subsection{Graph Interaction Module}
We design a graph structure for information learning and interaction between different modalities in GIM, which can improve the quality of fusion results. Furthermore, it enables images to contain more high-level information, so that the down-stream tasks ($ i.e. $, object detection, image segmentation) also perform well. The middle of Fig.~\ref{network} shows the specific workflow of the GIM. 

As the infrared branch an example, we provide the former features $ {\rm f}_i^{ir} $ with different scales acting as pioneer factors to the GIM for graph generation. Note that the GIM implements three loops of graph structures with three nodes in each branch to balance the performance of fusion results and operational efficiency. Detailed ablation experiments are conducted in Section.~\ref{Ablation Analysis}. In the process of creating graphs, we connect nodes of different scales from the same modality and nodes of the same scale from different modalities concurrently. The interactive way can restrict information imbalance while enhancing the representation of each input in fused images. After obtaining a graph, nodes constitute a corresponding leader node $ {\rm g}_i^{ir} $ to guide information delivery for the latter graph. Owing to the assistance of leader nodes, the GIM can resist information loss, improving the capability of feature learning. The leader nodes $ {\rm g}_1^{ir} $, $ {\rm g}_2^{ir} $ and $ {\rm g}_3^{ir} $ are finally mixed together to achieve the graph feature $ {\rm g}_{ir} $.

\subsubsection{Node and Edge Generation}

\begin{figure}
	\centering
	\includegraphics[width=0.48\textwidth]{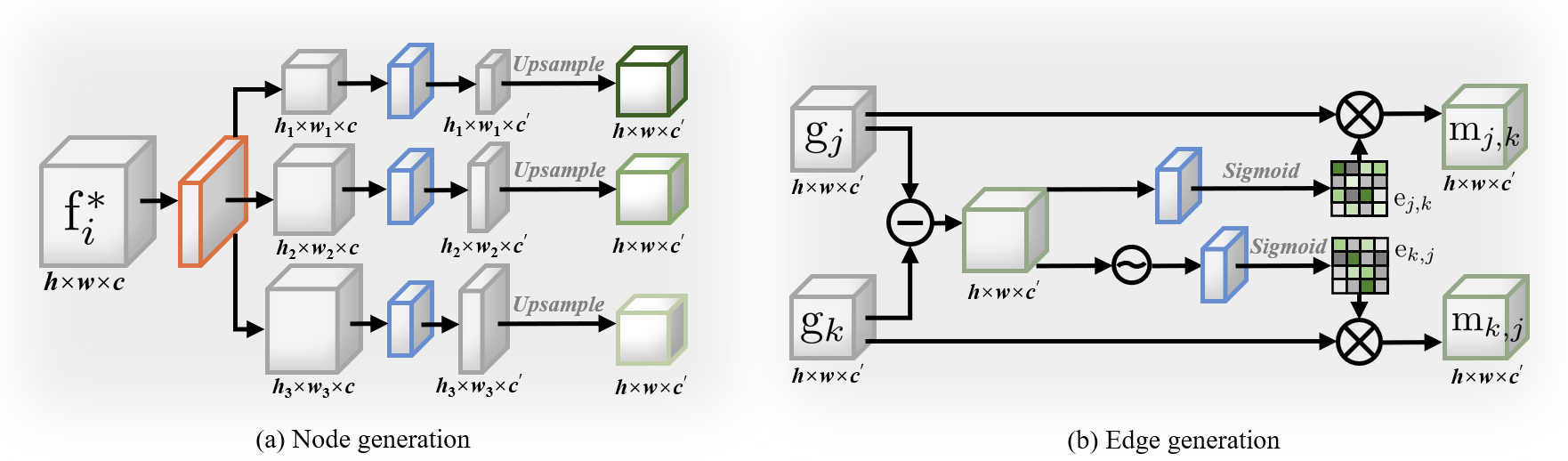}
	\caption{Specific illustration of (a) node generation and (b) edge generation.}
	\label{node-edge-generation}	
\end{figure}

Aimed at ensuring the diversity of features, we divide them into nodes of different scales through the pyramid pooling module (PPM) \cite{zhao2017pyramid}. Fig.~\ref{node-edge-generation} (a) describes the detailed process of node generation. We employ pyramid pooling, convolution, and upsample operations to split $ {\rm f}^{*}_i $ into multiple scales to obtain the nodes in the graph, respectively. Note that the nodes and $ {\rm f}^{*}_i $ keep consistent except for the number of channels. This process can be proved as follow:
\begin{equation}
	({\rm g}_i^*)_o = Up\Big( Conv\big( \mathcal{P}({\rm f}^{*}_i)\big)\Big),
\end{equation} 
where $ ({\rm g}_i^*)_o $ represents the $ o $-th ($ j \in \{1, 2, 3\}$) node of the $ i $-th ($ i \in \{1, 2, 3\}$) graph in * (infrared/visible) modality. $ Up $ and $ \mathcal{P} $ denote the upsample and pyramid pooling operations.

The production of edges in Fig.~\ref{node-edge-generation} (b) also stands an essential role of the graph generation, which carries the information transmission between nodes. We build edges in different-scale nodes from the same modality. Distinctively, nodes with the same scale from different modalities are linked for learning more semantic-level relations. The edge generation in $ {\rm g}_{j} $ and $ {\rm g}_{k} $ is bidirectional and defined as:
\begin{equation}
	{\rm e}_{j,k} = Conv({\rm g}_{j} - {\rm g}_{k}),
\end{equation}
\begin{equation}
	{\rm e}_{k,j} = Conv\big( \mathcal{N}({\rm g}_{j} - {\rm g}_{k})\big),
\end{equation}
where $ \mathcal{N} $ means the negation operation. $ {\rm e}_{j,k} $ ($ {\rm e}_{k,j} $) is the edge embedded from ${\rm g}_{j} $ ($ {\rm g}_{k} $) to ${\rm g}_{k} $ (${\rm g}_{j} $). In addition, the message passing $ {\rm m}_{j, k} $ can be formulated as:
\begin{equation}
	{\rm m}_{j,k} = Sigmoid({\rm e}_{j,k}) \cdot {\rm g}_{j},
\end{equation}
where $ Sigmoid $ denotes the Sigmoid operation.

\subsubsection{Leader Node and Information Delivery} 
\begin{figure}
	\centering
	\includegraphics[width=0.48\textwidth]{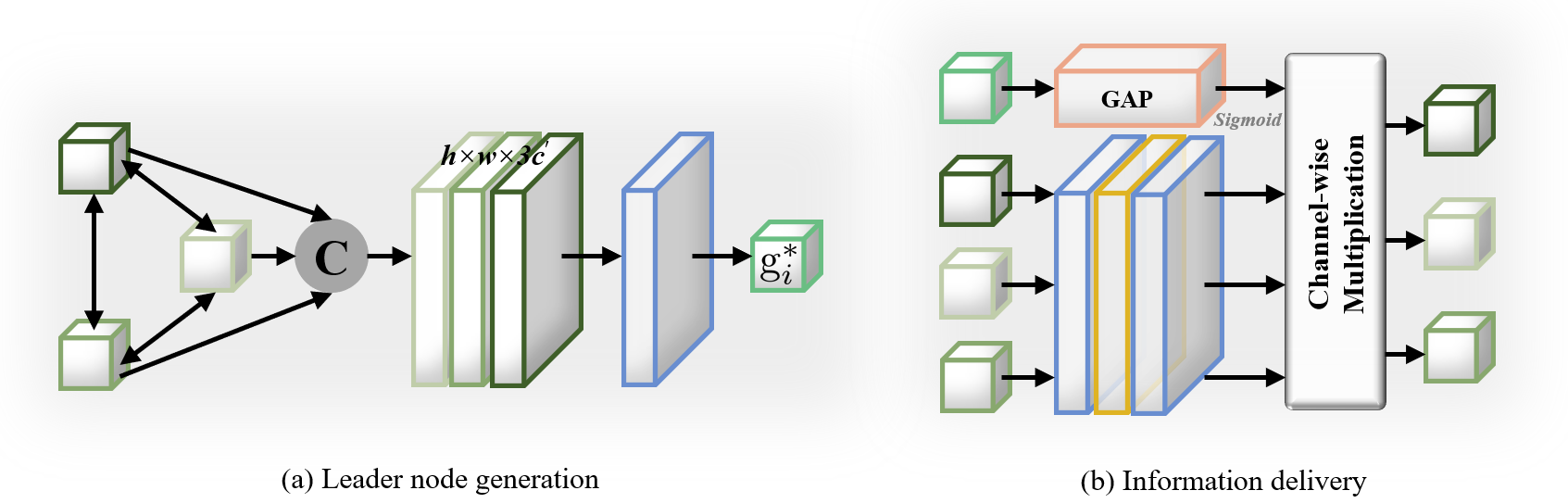}
	\caption{Specific illustration of (a) leader node generation and (b) information delivery.}
	\label{leader-node}	
\end{figure}

In Fig.~\ref{leader-node} (a), the introduction of leader nodes makes the delivery of semantic information between nodes in the graph more effectively, which can be represented as follow:
\begin{equation}
	{\rm g}_i^* = Conv\Big( Concat\big( ({\rm g}_i^*)_1, ({\rm g}_i^*)_2, ({\rm g}_i^*)_3\big)\Big)
\end{equation}

In the process of information delivery as shown in Fig.~\ref{leader-node} (b), the leader node generates the corresponding leader weight by the GAP and Sigmoid operation. After three former nodes pass through the convolutions, we multiply them with the leader weight in channel domain. Finally, the extracted multi-level features are propagated into the latter nodes, which can embody both details and targets clearly in fused images.

\subsection{Loss Function}
\label{Loss Function}
To guarantee that more meaningful information can be learned during the training phase, we introduce three varieties of loss functions, $ i.e. $, the pixel loss $ \mathcal{L}_{\rm MSE} $, the edge loss $ \mathcal{L}_{\rm edge} $ and the structure loss $ \mathcal{L}_{\rm SSIM} $. The combined $ \mathcal{L}_{\rm total} $ can be shown as follow:
\begin{equation}
	\mathcal{L}_{\rm total} = \mathcal{L}_{\rm MSE} + \alpha\mathcal{L}_{\rm edge} + \beta\mathcal{L}_{\rm SSIM},
\end{equation}
where $ \alpha $ and $ \beta $ are preset hyperparameters with the value of 10 and 0.5. Specifically, mean squared error (MSE) can measure the pixel intensity between source images and the fusion result. Note that we conduct weighted average to source images before calculating. It can be defined as:
\begin{equation}
	\mathcal{L}_{\rm MSE} = {\rm MSE}\big( ({\rm I}_{ir} + {\rm I}_{vis})/2, {\rm I}_f\big),
\end{equation}
where $ {\rm I}_{ir} $ and $ {\rm I}_{vis} $ mean infrared and visible images, respectively. To highlight edge details, $ \mathcal{L}_{\rm edge} $ selects the infrared/visible image with a larger gradient value to achieve the edge gradient:
\begin{equation}
	\mathcal{L}_{\rm edge} = \parallel \bigtriangledown{\rm I}_f - max(\bigtriangledown{\rm I}_{ir}, \bigtriangledown{\rm I}_{vis}) \parallel_{1}^{2},
\end{equation}
where $ \bigtriangledown $ is the gradient operator and $ \parallel \cdot \parallel_{1} $ is the $ l_1 $-norm. Besides, structural similarity index measure (SSIM) \cite{wang2004image} can calculate the similarity between source images and the fusion image, which is expressed as follow:
\begin{equation}
	\mathcal{L}_{\rm SSIM} = \big(1 - {\rm SSIM}({\rm I}_f, {\rm I}_{ir})\big) + \big(1 - {\rm SSIM}({\rm I}_f, {\rm I}_{vis})\big).
\end{equation}
With the help of the above loss function, the pixel and structural level information can be fully retained, which makes the fusion and down-stream results perform well.

\section{Experiments}
In this section, we first introduce the experimental setup, comparison approaches and dataset selection. Then, we analyze the fusion, detection, and segmentation results separately to verify the superiority of our proposed method. Furthermore, ablation experiments are mentioned to demonstrate the effectiveness of the proposed modules.

\subsection{Experimental Implementation}
In the training phase, we choose Adam optimizer to adjust the training parameters, where the stride and bitch size are set to 8 and 2. The initial learning rate of the network is $ 1{\rm e}^{-3} $ with a decay rate of $ 2{\rm e}^{-4} $. The total epoch is 100. In the loss function, the hyperparameters $ \alpha $ and $ \beta $ are set to 10 and 0.5, respectively. The selection of training datasets is presented in Section.~\ref{Comparison Approaches and Dataset Selection}. All experiments are implemented on an NVIDIA GeForce 3070Ti GPU with PyTorch framework.

\subsection{Dataset Selection and Comparison Approaches}
\label{Comparison Approaches and Dataset Selection}

The TNO \cite{toet2017tno}, $ {\rm M}^3$FD \cite{liu2022target} and MFNet \cite{ha2017mfnet} datasets contain plenty of infrared and visible image pairs. Moreover, the $ {\rm M}^3$FD and MFNet datasets also have image pairs that have been labeled for detection and segmentation. Before training, we combine 15 TNO pairs, 150 $ {\rm M}^3$FD pairs and 1083 MFNet pairs as the training set of our IGNet. The testing set consists of 10 TNO pairs, 150 $ {\rm M}^3$FD pairs, and 361 MFNet pairs. Note that the division of the TNO and $ {\rm M}^3$FD datasets is random, the MFNet dataset is based on \cite{tang2022image}.

We select seven state-of-the-art methods including DIDFuse \cite{zhao2020didfuse}, U2Fusion \cite{xu2020u2fusion}, SDNet \cite{zhang2021sdnet}, TarDAL \cite{liu2022target}, UMFusion \cite{wang2022unsupervised}, DeFusion \cite{liang2022fusion} and ReCoNet \cite{huang2022reconet}, to compare with our proposed IGNet in qualitative and quantitative results. During the fusion task, we apply six evaluation metrics, $ i.e. $, entropy (EN), visual information fidelity (VIF) \cite{han2013new}, average gradient (AG), correlation coefficient (CC) \cite{shah2013multifocus}, the sum of the correlations of differences (SCD) \cite{aslantas2015new} and edge-based
similarity measure ($ \rm{Q_{ab/f}}$) \cite{xydeas2000objective}, for objective estimation. Larger values of the above-mentioned metrics mean the  image quality performs better.

In the detection task, 4200 pairs of labeled images are employed as training, validation, and testing sets in a ratio of 8:1:1. The labels are marked into six categories, $ i.e. $, people, bus, car, motorcycle, truck, and lamp. A mainstream detector, YOLOv5 \cite{redmon2016you}, is conducted for detection. We set the optimizer, learning rate, epoch, and batch size as SGD optimizer, $ 1{\rm e}^{-2} $, 400, and 8, respectively. The mAP@.5 is measured for quantitative comparison. Moreover, we utilize DeepLabV3+ \cite{chen2018encoder} to segment fusion results, which choose the MFNet dataset as training and testing sets. There are nine labels in the sets, including background, car, person, bike, curve, car stop, guardrail, color cone, and bump. The training epoch and bitch size are set as 300 and 8, while other parameters keep the same as in the original experiment. The mIoU is selected for objective evaluation. In summary, we realize fusion images of each comparison approach to retain down-stream tasks, then analyze their corresponding performance.

\subsection{Analysis for Fusion Results}

\begin{figure*}
	\centering
	\includegraphics[width=0.98\textwidth]{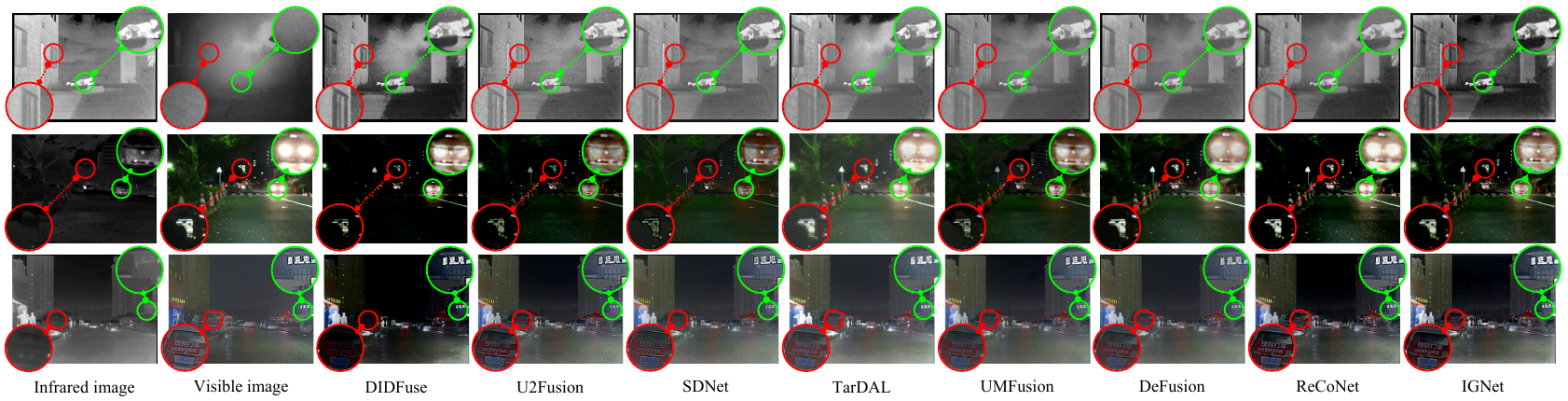}
	\caption{Visual comparisons of different approaches on TNO, MFNet and $ \bf{M^3}$FD datasets. Our proposed IGNet can achieve notable targets and fine background details. The enlarged red and green circles are detailed patches of fusion results.}
	\label{fusion}	
\end{figure*}

\begin{table*}[]
	\centering
	\caption{Quantitative comparisons of our IGNet with seven state-of-the-art methods on TNO, MFNet and $ \rm{M^3}$FD datasets. Optimal and suboptimal results are bolded in red and blue, respectively.}
	\label{quantitative}
	\renewcommand\tabcolsep{2pt}
	\begin{tabular}{c|cccccc|cccccc|cccccc}
		\toprule
		& \multicolumn{6}{c|}{\textbf{Dataset:TNO}}                                                                                                                                                                                                     & \multicolumn{6}{c|}{\textbf{Dataset:MFNet}}                                                                                                                                                                                                   & \multicolumn{6}{c}{\textbf{Dataset:$\textbf M^3$FD}} 
		\\
		\multirow{-2}{*}{\textbf{Method}} & EN & VIF & AG & CC & SCD & $ \rm{Q_{ab/f}}$ & EN & VIF & AG & CC & SCD & $ \rm{Q_{ab/f}}$ & EN & VIF & AG & CC & SCD & $ \rm{Q_{ab/f}}$ 
		\\ 
		\midrule
		DIDFuse & 7.066 & 0.738 & {\color[HTML]{0000FF} \textbf{5.150}} & 0.503 & 1.726 & 0.413 & 2.695 & 0.277 & 2.005 & 0.526 & 1.007 & 0.176 & 7.108 & {\color[HTML]{0000FF} \textbf{0.879}} & {\color[HTML]{0000FF} \textbf{5.663}} & 0.558 & 1.666 & 0.482 
		\\
		U2Fusion & 6.844 & 0.663 & 5.062 & 0.242 & {\color[HTML]{0000FF} \textbf{1.739}} & 0.444 & 4.612 & 0.503 & 2.899 & 0.627 & 1.262 & 0.364 & 7.090 & 0.831 & 5.546 & {\color[HTML]{0000FF} \textbf{0.569}} & {\color[HTML]{0000FF} \textbf{1.753}} & 0.524 
		\\
		SDNet & 6.682 & 0.661 & 5.059 & 0.501 & 1.562 & {\color[HTML]{0000FF} \textbf{0.450}} & 5.428 & 0.474 & 3.054 & {\color[HTML]{0000FF} \textbf{0.642}} & 1.111 & 0.410 & 7.013 & 0.729 & 5.514 & 0.500 & 1.544 & {\color[HTML]{0000FF} \textbf{0.525}} 
		\\
		TarDAL & {\color[HTML]{FF0000} \textbf{7.163}} & {\color[HTML]{FF0000} \textbf{0.800}} & 4.789 & 0.484 & 1.670 & 0.412 & {\color[HTML]{FF0000} \textbf{6.478}} & 0.699 & {\color[HTML]{0000FF} \textbf{3.140}} & 0.628 & {\color[HTML]{0000FF} \textbf{1.526}} & 0.420 & {\color[HTML]{0000FF} \textbf{7.126}} & 0.812 & 4.140 & 0.510 & 1.450 & 0.407                                 
		\\
		UMFusion  & 6.699 & 0.673& 3.710 &{\color[HTML]{0000FF} \textbf{0.516}} & 1.677 & 0.409 & 5.761 & 0.488 & 2.442 & 0.597 & 1.077 & 0.299 & 6.881 & 0.771 & 3.420 & 0.546 & 1.618 & 0.470                                 
		\\
		DeFusion & 6.724 & 0.712 & 2.996 & 0.493 & 1.592 & 0.325  & 5.950 & {\color[HTML]{0000FF} \textbf{0.759}} & 2.855 & 0.589 & 1.339 & {\color[HTML]{0000FF} \textbf{0.471}} & 6.634 & 0.740 & 3.027 & 0.513 & 1.366 & 0.412  
		\\
		ReCoNet & 6.682 & 0.728 & 3.674 & 0.481 & 1.732 & 0.340 & 3.894 & 0.544 & 3.105 & 0.544 & 1.190 & 0.365 & 6.740 & 0.867 & 4.557 & 0.515 & 1.495 & 0.499                                 
		\\
		IGNet & {\color[HTML]{0000FF} \textbf{7.099}} & {\color[HTML]{0000FF} \textbf{0.764}} & {\color[HTML]{FF0000} \textbf{5.247}} & {\color[HTML]{FF0000} \textbf{0.521}} & {\color[HTML]{FF0000} \textbf{1.756}} & {\color[HTML]{FF0000} \textbf{0.459}} & {\color[HTML]{0000FF} \textbf{6.124}} & {\color[HTML]{FF0000} \textbf{0.762}} & {\color[HTML]{FF0000} \textbf{3.290}} & {\color[HTML]{FF0000} \textbf{0.655}} & {\color[HTML]{FF0000} \textbf{1.562}} & {\color[HTML]{FF0000} \textbf{0.485}} & {\color[HTML]{FF0000} \textbf{7.140}} & {\color[HTML]{FF0000} \textbf{0.882}} & {\color[HTML]{0000FF} \textbf{5.615}} & {\color[HTML]{FF0000} \textbf{0.575}} & {\color[HTML]{FF0000} \textbf{1.762}} & {\color[HTML]{FF0000} \textbf{0.539}} 
		\\ 
		\bottomrule
	\end{tabular}
\end{table*}

\begin{figure*}
	\centering
	\includegraphics[width=0.98\textwidth]{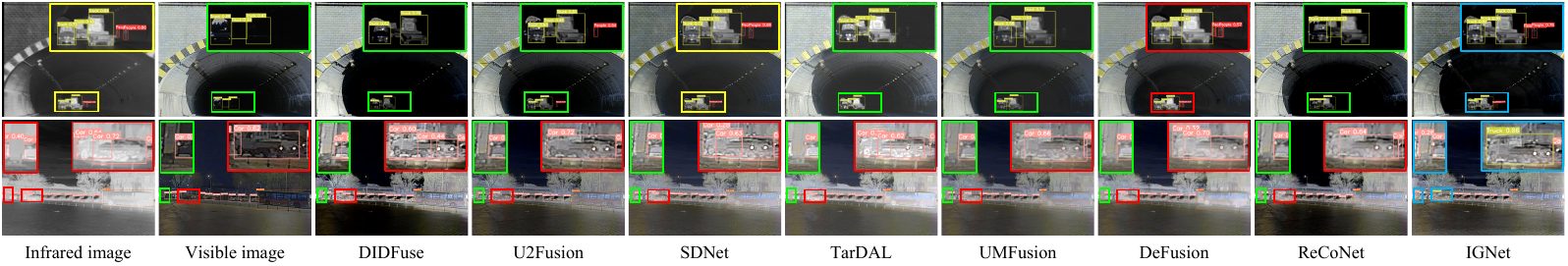}
	\caption{Detection visual comparisons of different fusion images on $ \bf{M^3}$FD dataset. Our proposed IGNet can generate high-confidence detection results with visually appealing performance. The red, green, and yellow regions represent error, missing, and low-confidence detection, respectively. The blue areas denote our outstanding details.}
	\label{detection}	
\end{figure*}

\begin{table}[]
	\centering
	\caption{Detection quantitative comparisons of our IGNet with seven state-of-the-art methods on $ \rm{M^3}$FD dataset. Optimal and suboptimal results are bolded in red and blue, respectively.}
	\label{Detection}
	\renewcommand\tabcolsep{1.8pt}
	\begin{tabular}{c|cccccc|c}
		\toprule
		& \multicolumn{6}{c|}{\textbf{AP@.5}}                                                                                                                                                                                                           & \multicolumn{1}{l}{}                                  \\
		\multirow{-2}{*}{\textbf{Method}} & People & Bus & Car & Motor & Truck & Lamp & \multicolumn{1}{l}{\multirow{-2}{*}{\textbf{mAP@.5}}} 
		\\ 
		\midrule
		Infrared & 0.807 & 0.782 & 0.888 & 0.640 & 0.652 & 0.703 & 0.745 
		\\
		Visible & 0.708 & 0.780 & 0.911 & 0.702 & 0.697 & 0.865 & 0.777 
		\\
		DIDFuse & 0.800 & 0.798 & 0.924 & 0.681 & 0.692 & 0.843 &0.790 
		\\
		U2Fusion & 0.793 & 0.785 & 0.916 & 0.663 & 0.710 & 0.872 & 0.789 
		\\
		SDNet & 0.790 & 0.811 & 0.920 & 0.670 & 0.689 & 0.838  & 0.786 
		\\
		TarDAL & {\color[HTML]{FF0000} \textbf{0.817}} & 0.815 & {\color[HTML]{FF0000} \textbf{0.948}} & 0.696 & 0.687 & 0.873 & {\color[HTML]{0000FF} \textbf{0.806}}                 
		\\
		UMFusion & 0.790 & 0.783 & 0.920 & {\color[HTML]{0000FF} \textbf{0.728}} & 0.691 & 0.847 & 0.793 
		\\
		DeFusion & 0.805 & {\color[HTML]{FF0000} \textbf{0.827}} &0.921  & 0.689 & {\color[HTML]{0000FF} \textbf{0.714}} & {\color[HTML]{FF0000} \textbf{0.876}} & 0.805 
		\\
		ReCoNet & 0.792 & 0.784 & 0.915 & 0.693 & 0.698 & {\color[HTML]{0000FF} \textbf{0.873}} & 0.792 
		\\
		IGNet & {\color[HTML]{0000FF} \textbf{0.816}} & {\color[HTML]{0000FF} \textbf{0.824}} & {\color[HTML]{0000FF} \textbf{0.928}} & {\color[HTML]{FF0000} \textbf{0.730}} & {\color[HTML]{FF0000} \textbf{0.721}} & 0.869 & {\color[HTML]{FF0000}\textbf{0.815}}               
		\\ 
		\bottomrule
	\end{tabular}
\end{table}

\subsubsection{Qualitative Analysis} 
We depict qualitative results on TNO, MFNet, and $ {\rm M}^3$FD datasets in Fig.~\ref{fusion}. Obviously, our results outperform other state-of-the-art methods. For instance, targets and surrounding scenes obscured by smoke can be clearly displayed on the TNO dataset. In the second illustration, TarDAL and ReCoNet occur over-exposed regions, while U2Fusion, SDNet and UMFusion remain low-contrast performance. Although DIDFusion can highlight luminance information ($ e.g. $, car lights), its background abandons many texture details, which is unfriendly to HVS. In addition, benefiting from the cooperation of GNN, blur artifacts can be effectively mitigated as shown in the green enlarged patch of the third row.

\subsubsection{Quantitative Analysis}

In Table.~\ref{quantitative}, we enumerate the mean scores for the six metrics in the three testing sets. From an overall perspective, the quantitative results of our method stand in the lead position. Specifically, CC and SCD achieve the highest scores, which indicates the mutual connection between our fusion images and source inputs is the tightest. The highest value of $ \rm{Q_{ab/f}}$ reflects that the edge contours of targets can be well represented. Moreover, the higher performance of EN and AG demonstrates that a large amount of information is preserved in our fusion results. Since our approach pays greater emphasis on information delivery, the VIF value also keeps at a higher level.

\subsection{Analysis for Detection Results}

\subsubsection{Qualitative Analysis} 
As shown in Fig.~\ref{detection}, the disturbance of environmental factors causes the detection results of single-modal images to be generally weaker than those of fusion results. However, the sensitivity of different fusion results to detection is also varied. In the first row of examples, SDNet and DeFusion present significantly low confidence and error detection regions, which may mislead observers. Moreover, "Truck" is detected as "Car" in the second set, while missing detection of cars in the corner also occurs. As a representative, our fusion results contain rich advanced features, so that the corresponding detection results can avoid the above phenomena. We can also notice that our detection results achieve high-confidence scores on all labeled categories.

\subsubsection{Quantitative Analysis}
Table.~\ref{Detection} exhibits the AP@.5 of each label and matching total mAP@.5 measured by detection results of fused images, which can obtain higher indicators than single infrared or visible images. Under the comparison of fusion methods, our proposed IGNet performs 2.59\% higher than others in detection. It is worth noting that IGNet can not only achieve excellent detection results but also take into account the quality of fusion images.

\subsection{Analysis for Segmentation Results}

\subsubsection{Qualitative Analysis} 
\begin{figure*}
	\centering
	\includegraphics[width=0.98\textwidth]{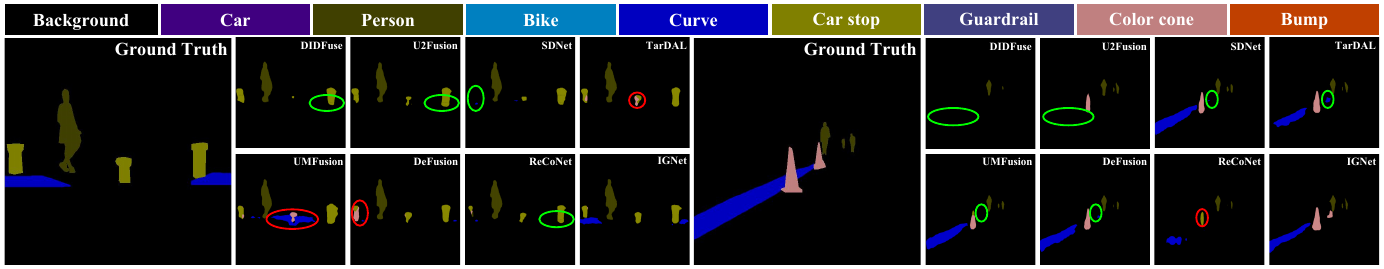}
	\caption{Segmentation visual comparisons of different fusion images on MFNet dataset. Our proposed IGNet can get the most accurate segmentation results compared to the ground truth. The red and green regions represent the error and missing segmentation, respectively.}
	\label{segmentation}	
\end{figure*}

Visual results of the segmentation on the MFNet dataset are presented in Fig.~\ref{segmentation}. Similarly, we also employ fusion results of each method as the input to obtain segmentation results. Due to less semantic information contained in images, DIDFuse, U2Fusion, SDNet, and ReCoNet appear some missing segmentation areas in the first sample. In addition to the segmentation results of IGNet, the "Color cone" in the second example cannot be accurately segmented. It is appropriate to mention that our proposed method can exploit cross-modality interaction features to efficiently segment the contours of labeled objects.

\subsubsection{Quantitative Analysis}
Table.~\ref{Segmentation} depicts the segmentation quantitative metric IoU for different categories, which presents IGNet outperforms other fusion methods in the segmentation task. Compared with the second-ranked method, our method improves mIoU in the ratio of 4.87\%. For some infrared-sensitiveness labels, $ e.g. $, person, higher scores indicated that our method can more easily highlight thermal targets. Due to the high fidelity of fused images, the IoU of some visually appealing labels, $ e.g. $, car and bike, still keeps high performance. Note that the proposed IGNet can also generate vivid fusion images while achieving accurate segmentation results.

\subsection{Ablation Experiments}
\label{Ablation Analysis}

\subsubsection{Study on Modules}
\begin{figure}
	\centering
	\includegraphics[width=0.48\textwidth]{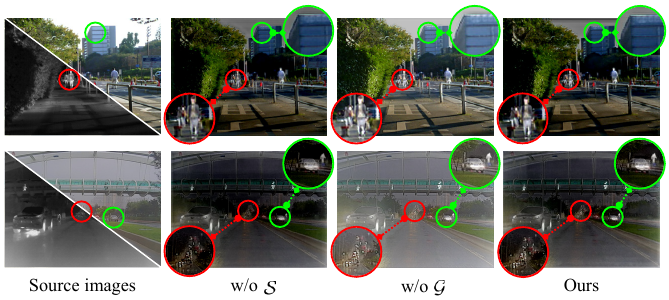}
	\caption{Visual ablation comparisons of the SSM ($ \mathcal{S} $) and GIM ($ \mathcal{G} $) about fusion. The enlarged red and green circles are detailed patches of fusion results.}
	\label{ablation-model-fusion}	
\end{figure}

\begin{figure}
	\centering
	\includegraphics[width=0.48\textwidth]{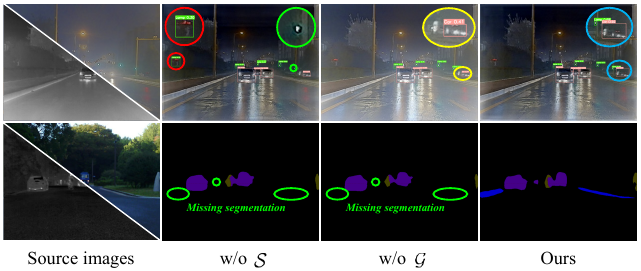}
	\caption{Visual ablation comparisons of the SSM ($ \mathcal{S} $) and GIM ($ \mathcal{G} $) about down-stream tasks. The detailed regions are marked.}
	\label{ablation-detseg}	
\end{figure}

The proposed SSM and GIM play a key role in improving the fusion effect. It is obvious that fusion results perform poorly in luminance without SSM as shown in Fig.~\ref{ablation-model-fusion}. Also, the cross-modality features of infrared and visible branches can not interact with each other without the decoration of the GIM, which causes the low contrast and halo artifacts of images. Furthermore, Fig.~\ref{ablation-detseg} reports the results of down-stream tasks. Due to the abundant semantic information extracted by the proposed module, the full modal can simultaneously obtain high-confidence detection and accurate segmentation results. The quantitative comparisons are depicted in Table.~\ref{Ablation quantitative}. It is not difficult to prove that the utilization of our proposed modules can bridge fusion and downstream tasks with a mutually beneficial situation.

\begin{table}[]
	\centering
	\caption{Segmentation quantitative comparisons of our IGNet with seven state-of-the-art methods on MFNet dataset. Optimal and suboptimal results are bolded in red and blue, respectively.}
	\label{Segmentation}
	\renewcommand\tabcolsep{1.3pt}
	\scalebox{0.83}{
		\begin{tabular}{c|ccccccccc|c}
			\toprule
			& \multicolumn{9}{c|}{\textbf{IoU}} &                                       
			\\
			\multirow{-2}{*}{\textbf{Method}} & \multicolumn{1}{l}{Bac} & Car & Per & Bik & Cur & C S & Gua & C C & Bum & \multirow{-2}{*}{\textbf{mIoU}}       
			\\ 
			\midrule
			Infrared & 0.821 & 0.663 & 0.592 & 0.513 & 0.347 & 0.398 & 0.422 & 0.414  & 0.479 & 0.516
			\\
			Visible & 0.899 & 0.774 & 0.482 & 0.586 & 0.372 & 0.517 & 0.451 & 0.432 & 0.506 & 0.558 
			\\
			DIDFuse & 0.971 & 0.790 & 0.582 & 0.599 & 0.358 & {\color[HTML]{0000FF} \textbf{0.526}} & {\color[HTML]{0000FF} \textbf{0.619}} & 0.442 & 0.557 & 0.604                                 
			\\
			U2Fusion & 0.974 & 0.817 & {\color[HTML]{0000FF} \textbf{0.631}} & {\color[HTML]{0000FF} \textbf{0.625}} & 0.408 & 0.523 & 0.520 & 0.448 & {\color[HTML]{FF0000} \textbf{0.593}} & {\color[HTML]{0000FF} \textbf{0.615}} 
			\\
			SDNet & 0.973 & 0.782 & 0.614 & 0.618 & 0.361 & 0.500 & 0.527 & 0.425 & 0.527 & 0.591 
			\\
			TarDAL & 0.970 & 0.795 & 0.563 & 0.591 & 0.342 & 0.497 & 0.553 & 0.425 & 0.538 & 0.586
			\\
			UMFusion & 0.972 & 0.787 & 0.607 & 0.616 & 0.364 & 0.493 & 0.479 & 0.447 & 0.485 & 0.583 
			\\
			DeFusion & {\color[HTML]{0000FF} \textbf{0.975}} & {\color[HTML]{0000FF} \textbf{0.820}} & 0.609 & 0.623 & 0.401 & 0.488 & 0.482 & 0.471 & 0.548 & 0.601                                 
			\\
			ReCoNet & 0.973 & 0.813 & 0.598 & 0.610 & {\color[HTML]{0000FF} \textbf{0.413}} & 0.519 & 0.544 & {\color[HTML]{0000FF} \textbf{0.476}} & 0.552 & 0.610                                 
			\\
			IGNet & {\color[HTML]{FF0000} \textbf{0.976}} & {\color[HTML]{FF0000} \textbf{0.838}} & {\color[HTML]{FF0000} \textbf{0.639}} & {\color[HTML]{FF0000} \textbf{0.667}} & {\color[HTML]{FF0000} \textbf{0.435}} & {\color[HTML]{FF0000} \textbf{0.532}} & {\color[HTML]{FF0000} \textbf{0.626}} & {\color[HTML]{FF0000} \textbf{0.511}} & {\color[HTML]{0000FF} \textbf{0.586}} & {\color[HTML]{FF0000} \textbf{0.645}} 
			\\ 
			\bottomrule
	\end{tabular}}
\end{table}

\begin{table*}[]
	\centering
	\caption{Quantitative ablation results of modules on two different datasets. Optimal and suboptimal results are bolded in red and blue, respectively.}
	\label{Ablation quantitative}
	\begin{tabular}{ccc|ccccccc|ccccccc}
		\toprule
		& & & \multicolumn{7}{c|}{\textbf{Dataset:$\textbf M^3$FD}}  & \multicolumn{7}{c}{\textbf{Dataset:MFNet}}
		\\
		\multirow{-2}{*}{\textbf{Model}} & \multirow{-2}{*}{$ \mathcal{S} $} & \multirow{-2}{*}{$ \mathcal{G} $} & EN & VIF & AG & CC & SCD & $ \rm{Q_{ab/f}} $ & mAP@.5 & EN & VIF & AG & CC & SCD & $ \rm{Q_{ab/f}} $ & MIoU 
		\\ 
		\midrule
		M1 & \usym{2717} & \usym{2714} & {\color[HTML]{0000FF} \textbf{7.052}} & {\color[HTML]{0000FF} \textbf{0.873}} & {\color[HTML]{0000FF} \textbf{5.608}} & {\color[HTML]{0000FF} \textbf{0.570}} & {\color[HTML]{0000FF} \textbf{1.699}} & 0.504 & {\color[HTML]{0000FF} \textbf{0.801}} & {\color[HTML]{0000FF} \textbf{6.045}} & {\color[HTML]{0000FF} \textbf{0.751}} & {\color[HTML]{0000FF} \textbf{3.187}} & 0.633 & 1.549 & {\color[HTML]{0000FF} \textbf{0.470}} & {\color[HTML]{0000FF} \textbf{0.624}} 
		\\
		M2 & \usym{2714} & \usym{2717} & 7.009 & 0.735 & 5.536 & 0.507 & 1.574 & {\color[HTML]{0000FF} \textbf{0.515}} & 0.791 & 5.443 & 0.498 & 3.091 & {\color[HTML]{0000FF} \textbf{0.642}} & {\color[HTML]{0000FF} \textbf{1.551}} & 0.426 & 0.587 
		\\
		M3 & \usym{2714} & \usym{2714} & {\color[HTML]{FF0000} \textbf{7.140}} & {\color[HTML]{FF0000} \textbf{0.882}} & {\color[HTML]{FF0000} \textbf{5.615}} & {\color[HTML]{FF0000} \textbf{0.575}} & {\color[HTML]{FF0000} \textbf{1.762}} & {\color[HTML]{FF0000} \textbf{0.539}} & {\color[HTML]{FF0000} \textbf{0.815}} & {\color[HTML]{FF0000} \textbf{6.124}} & {\color[HTML]{FF0000} \textbf{0.762}} & {\color[HTML]{FF0000} \textbf{3.290}} & {\color[HTML]{FF0000} \textbf{0.655}} & {\color[HTML]{FF0000} \textbf{1.562}} & {\color[HTML]{FF0000} \textbf{0.485}} & {\color[HTML]{FF0000} \textbf{0.645}} 
		\\ 
		\bottomrule
	\end{tabular}
\end{table*}

\begin{table}
	\centering
	\caption{Quantitative ablation results about the number of nodes (N) and loops(L) in the graph. Optimal and suboptimal results are bolded in red and blue, respectively.}
	\label{Ablation GNN}
	\renewcommand\tabcolsep{4pt}
	\begin{tabular}{ccc|cccccc}
		\toprule
		\textbf{Model} & N & L & EN & VIF & AG & CC & SCD  & $\rm{Q_{ab/f}}$                                  
		\\ 
		\midrule
		M1 & 1 & 3 & 6.032 & 0.751 & 3.289 & 0.647 & 1.553 & 0.480                                 
		\\
		M2 & 3 & 3 & {\color[HTML]{0000FF} \textbf{6.124}} & {\color[HTML]{0000FF} \textbf{0.762}} & {\color[HTML]{FF0000} \textbf{3.290}} & {\color[HTML]{0000FF} \textbf{0.655}} & {\color[HTML]{0000FF} \textbf{1.562}} & {\color[HTML]{0000FF} \textbf{0.485}} 
		\\
		M3 & 5 & 3 & {\color[HTML]{FF0000} \textbf{6.125}} & {\color[HTML]{0000FF} \textbf{0.762}} & {\color[HTML]{0000FF} \textbf{3.289}} & {\color[HTML]{FF0000} \textbf{0.657}} & {\color[HTML]{FF0000} \textbf{1.563}} & {\color[HTML]{0000FF} \textbf{0.485}} 
		\\ 
		\midrule
		M1 & 3 & 1 & 6.111 & 0.744 & 3.277 & 0.641 & 1.559 & 0.476                                 
		\\
		M2 & 3 & 3 & {\color[HTML]{0000FF} \textbf{6.124}} & {\color[HTML]{0000FF} \textbf{0.762}} & {\color[HTML]{0000FF} \textbf{3.290}} & {\color[HTML]{0000FF} \textbf{0.655}} & {\color[HTML]{0000FF} \textbf{1.562}} & {\color[HTML]{0000FF} \textbf{0.485}} 
		\\
		M3 & 3 & 5 & {\color[HTML]{0000FF} \textbf{6.124}} & {\color[HTML]{FF0000} \textbf{0.764}} & {\color[HTML]{FF0000} \textbf{0.291}} & {\color[HTML]{0000FF} \textbf{0.655}} & {\color[HTML]{FF0000} \textbf{1.564}} & {\color[HTML]{FF0000} \textbf{0.487}} 
		\\ 
		\bottomrule
	\end{tabular}
\end{table}

\subsubsection{Study on Leader Node}
\begin{figure}
	\centering
	\includegraphics[width=0.48\textwidth]{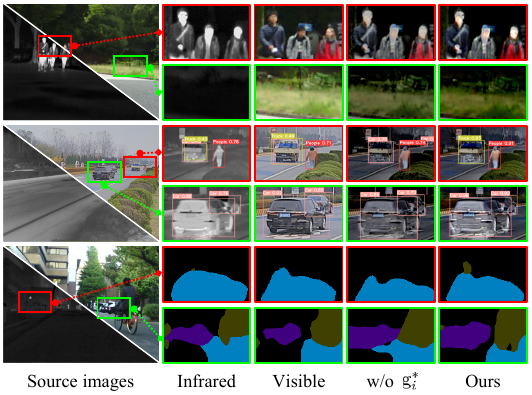}
	\caption{Visual ablation comparisons of the leader nodes ($ {\rm g}_i^* $) about fusion, detection and segmentation. With the help of leader nodes, the image details and down-stream results can perform more vividly and accurately. The enlarged red and green boxes are detailed patches of corresponding results.}
	\label{ablation-leader-node}	
\end{figure}

In order to avoid intermediate feature loss, we use leader nodes to guide the information delivery. Without the help of leader nodes, fused images often appear distorting in color. Meanwhile, some wrong regions may emerge in detection and segmentation results. In contrast, IGNet makes full use of feature maps delivered by the leader nodes inside graphs, enabling semantic information to be revealed in fused images. Fig.~\ref{ablation-leader-node} performs the superiority of our proposed method on two different datasets.

\subsubsection{Study on Parameters of Graph}
We select one, three and five nodes to conduct each graph structure, aiming at verifying how the number of nodes N influence results. Except for the number of nodes, other parameters remain unchanged. It can be seen from Table.~\ref{Ablation GNN} that when there is only a single node in a graph, the quantitative indicators perform undesirably. As the number increases to five, its performance is almost indistinguishable from our results (N = 3). However, the operating efficiency of the network will decrease with N rising. Considering this issue, we employ three nodes in each graph, which can balance the quality of images and inference speed. Similarly, the number of loop L are preset to three for a trade-off.

\section{Conclusion}

In this paper, an interactive cross-modality framework based on graph neural network was proposed for infrared and visible image fusion. We presented a graph interaction module to learn mutual features from different branches, which can emphasize outstanding textures in source images. Aiming at preventing information from missing, the leader nodes were proposed to guide the feature propagation between adjacent graphs. In addition, abundant semantic information was also extracted by our proposed method, thus we could achieve well-performance detection and segmentation results. Extensive experiments proved our method is advanced in IVIF and down-stream tasks.
 
In the future, we tend to bridge multi-modality fusion, target detection, and image segmentation in a unified framework. In other words, it is worth further exploiting how to generate a fusion image that can also perform well in detection and segmentation tasks.

\section*{Acknowledgments}
This work was supported by the National Key R\&D Program of China (2022ZD0117902) and by the National Natural Science Foundation of China (No. U20B2062).

\bibliographystyle{ACM-Reference-Format}
\bibliography{IGNet_arxiv}


\begin{thebibliography}{47}


\ifx \showCODEN    \undefined \def \showCODEN     #1{\unskip}     \fi
\ifx \showDOI      \undefined \def \showDOI       #1{#1}\fi
\ifx \showISBNx    \undefined \def \showISBNx     #1{\unskip}     \fi
\ifx \showISBNxiii \undefined \def \showISBNxiii  #1{\unskip}     \fi
\ifx \showISSN     \undefined \def \showISSN      #1{\unskip}     \fi
\ifx \showLCCN     \undefined \def \showLCCN      #1{\unskip}     \fi
\ifx \shownote     \undefined \def \shownote      #1{#1}          \fi
\ifx \showarticletitle \undefined \def \showarticletitle #1{#1}   \fi
\ifx \showURL      \undefined \def \showURL       {\relax}        \fi
\providecommand\bibfield[2]{#2}
\providecommand\bibinfo[2]{#2}
\providecommand\natexlab[1]{#1}
\providecommand\showeprint[2][]{arXiv:#2}

\bibitem[\protect\citeauthoryear{Aslantas and Bendes}{Aslantas and
  Bendes}{2015}]%
        {aslantas2015new}
\bibfield{author}{\bibinfo{person}{V Aslantas} {and} \bibinfo{person}{Emre
  Bendes}.} \bibinfo{year}{2015}\natexlab{}.
\newblock \showarticletitle{A new image quality metric for image fusion: The
  sum of the correlations of differences}.
\newblock \bibinfo{journal}{\emph{Aeu-international Journal of electronics and
  communications}} \bibinfo{volume}{69}, \bibinfo{number}{12}
  (\bibinfo{year}{2015}), \bibinfo{pages}{1890--1896}.
\newblock


\bibitem[\protect\citeauthoryear{Chen, Zhu, Papandreou, Schroff, and Adam}{Chen
  et~al\mbox{.}}{2018}]%
        {chen2018encoder}
\bibfield{author}{\bibinfo{person}{Liang-Chieh Chen}, \bibinfo{person}{Yukun
  Zhu}, \bibinfo{person}{George Papandreou}, \bibinfo{person}{Florian Schroff},
  {and} \bibinfo{person}{Hartwig Adam}.} \bibinfo{year}{2018}\natexlab{}.
\newblock \showarticletitle{Encoder-decoder with atrous separable convolution
  for semantic image segmentation}. In \bibinfo{booktitle}{\emph{Proceedings of
  the European conference on computer vision (ECCV)}}.
  \bibinfo{pages}{801--818}.
\newblock


\bibitem[\protect\citeauthoryear{Ha, Watanabe, Karasawa, Ushiku, and Harada}{Ha
  et~al\mbox{.}}{2017}]%
        {ha2017mfnet}
\bibfield{author}{\bibinfo{person}{Qishen Ha}, \bibinfo{person}{Kohei
  Watanabe}, \bibinfo{person}{Takumi Karasawa}, \bibinfo{person}{Yoshitaka
  Ushiku}, {and} \bibinfo{person}{Tatsuya Harada}.}
  \bibinfo{year}{2017}\natexlab{}.
\newblock \showarticletitle{MFNet: Towards real-time semantic segmentation for
  autonomous vehicles with multi-spectral scenes}. In
  \bibinfo{booktitle}{\emph{2017 IEEE/RSJ International Conference on
  Intelligent Robots and Systems (IROS)}}. IEEE, \bibinfo{pages}{5108--5115}.
\newblock


\bibitem[\protect\citeauthoryear{Han, Cai, Cao, and Xu}{Han
  et~al\mbox{.}}{2013}]%
        {han2013new}
\bibfield{author}{\bibinfo{person}{Yu Han}, \bibinfo{person}{Yunze Cai},
  \bibinfo{person}{Yin Cao}, {and} \bibinfo{person}{Xiaoming Xu}.}
  \bibinfo{year}{2013}\natexlab{}.
\newblock \showarticletitle{A new image fusion performance metric based on
  visual information fidelity}.
\newblock \bibinfo{journal}{\emph{Information fusion}} \bibinfo{volume}{14},
  \bibinfo{number}{2} (\bibinfo{year}{2013}), \bibinfo{pages}{127--135}.
\newblock


\bibitem[\protect\citeauthoryear{Hu, Shen, and Sun}{Hu et~al\mbox{.}}{2018}]%
        {hu2018squeeze}
\bibfield{author}{\bibinfo{person}{Jie Hu}, \bibinfo{person}{Li Shen}, {and}
  \bibinfo{person}{Gang Sun}.} \bibinfo{year}{2018}\natexlab{}.
\newblock \showarticletitle{Squeeze-and-excitation networks}. In
  \bibinfo{booktitle}{\emph{Proceedings of the IEEE conference on computer
  vision and pattern recognition}}. \bibinfo{pages}{7132--7141}.
\newblock


\bibitem[\protect\citeauthoryear{Huang, Lin, Zhang, Xu, Zheng, Mao, Qian, Peng,
  Zhou, Chen, et~al\mbox{.}}{Huang et~al\mbox{.}}{2021}]%
        {huang2021graph}
\bibfield{author}{\bibinfo{person}{Huimin Huang}, \bibinfo{person}{Lanfen Lin},
  \bibinfo{person}{Yue Zhang}, \bibinfo{person}{Yingying Xu},
  \bibinfo{person}{Jing Zheng}, \bibinfo{person}{XiongWei Mao},
  \bibinfo{person}{Xiaohan Qian}, \bibinfo{person}{Zhiyi Peng},
  \bibinfo{person}{Jianying Zhou}, \bibinfo{person}{Yen-Wei Chen},
  {et~al\mbox{.}}} \bibinfo{year}{2021}\natexlab{}.
\newblock \showarticletitle{Graph-bas3net: Boundary-aware semi-supervised
  segmentation network with bilateral graph convolution}. In
  \bibinfo{booktitle}{\emph{Proceedings of the IEEE/CVF International
  Conference on Computer Vision}}. \bibinfo{pages}{7386--7395}.
\newblock


\bibitem[\protect\citeauthoryear{Huang, Liu, Fan, Liu, Zhong, and Luo}{Huang
  et~al\mbox{.}}{2022}]%
        {huang2022reconet}
\bibfield{author}{\bibinfo{person}{Zhanbo Huang}, \bibinfo{person}{Jinyuan
  Liu}, \bibinfo{person}{Xin Fan}, \bibinfo{person}{Risheng Liu},
  \bibinfo{person}{Wei Zhong}, {and} \bibinfo{person}{Zhongxuan Luo}.}
  \bibinfo{year}{2022}\natexlab{}.
\newblock \showarticletitle{ReCoNet: Recurrent Correction Network for Fast and
  Efficient Multi-modality Image Fusion}. In \bibinfo{booktitle}{\emph{Computer
  Vision--ECCV 2022: 17th European Conference, Tel Aviv, Israel, October
  23--27, 2022, Proceedings, Part XVIII}}. Springer, \bibinfo{pages}{539--555}.
\newblock


\bibitem[\protect\citeauthoryear{Jiang, Zhang, Fan, and Liu}{Jiang
  et~al\mbox{.}}{2022}]%
        {jiang2022towards}
\bibfield{author}{\bibinfo{person}{Zhiying Jiang}, \bibinfo{person}{Zengxi
  Zhang}, \bibinfo{person}{Xin Fan}, {and} \bibinfo{person}{Risheng Liu}.}
  \bibinfo{year}{2022}\natexlab{}.
\newblock \showarticletitle{Towards all weather and unobstructed multi-spectral
  image stitching: Algorithm and benchmark}. In
  \bibinfo{booktitle}{\emph{Proceedings of the 30th ACM International
  Conference on Multimedia}}. \bibinfo{pages}{3783--3791}.
\newblock


\bibitem[\protect\citeauthoryear{Lei, Li, Liu, Zhou, Zhang, and Kasabov}{Lei
  et~al\mbox{.}}{2023}]%
        {lei2023galfusion}
\bibfield{author}{\bibinfo{person}{Jia Lei}, \bibinfo{person}{Jiawei Li},
  \bibinfo{person}{Jinyuan Liu}, \bibinfo{person}{Shihua Zhou},
  \bibinfo{person}{Qiang Zhang}, {and} \bibinfo{person}{Nikola~K Kasabov}.}
  \bibinfo{year}{2023}\natexlab{}.
\newblock \showarticletitle{GALFusion: Multi-exposure Image Fusion via a
  Global-local Aggregation Learning Network}.
\newblock \bibinfo{journal}{\emph{IEEE Transactions on Instrumentation and
  Measurement}} (\bibinfo{year}{2023}).
\newblock


\bibitem[\protect\citeauthoryear{Li and Wu}{Li and Wu}{2018}]%
        {li2018densefuse}
\bibfield{author}{\bibinfo{person}{Hui Li} {and} \bibinfo{person}{Xiao-Jun
  Wu}.} \bibinfo{year}{2018}\natexlab{}.
\newblock \showarticletitle{DenseFuse: A fusion approach to infrared and
  visible images}.
\newblock \bibinfo{journal}{\emph{IEEE Transactions on Image Processing}}
  \bibinfo{volume}{28}, \bibinfo{number}{5} (\bibinfo{year}{2018}),
  \bibinfo{pages}{2614--2623}.
\newblock


\bibitem[\protect\citeauthoryear{Li, Huo, Li, Wang, and Feng}{Li
  et~al\mbox{.}}{2020}]%
        {li2020attentionfgan}
\bibfield{author}{\bibinfo{person}{Jing Li}, \bibinfo{person}{Hongtao Huo},
  \bibinfo{person}{Chang Li}, \bibinfo{person}{Renhua Wang}, {and}
  \bibinfo{person}{Qi Feng}.} \bibinfo{year}{2020}\natexlab{}.
\newblock \showarticletitle{AttentionFGAN: Infrared and visible image fusion
  using attention-based generative adversarial networks}.
\newblock \bibinfo{journal}{\emph{IEEE Transactions on Multimedia}}
  \bibinfo{volume}{23} (\bibinfo{year}{2020}), \bibinfo{pages}{1383--1396}.
\newblock


\bibitem[\protect\citeauthoryear{Li, Liu, Zhou, Zhang, and Kasabov}{Li
  et~al\mbox{.}}{2022}]%
        {li2022learning}
\bibfield{author}{\bibinfo{person}{Jiawei Li}, \bibinfo{person}{Jinyuan Liu},
  \bibinfo{person}{Shihua Zhou}, \bibinfo{person}{Qiang Zhang}, {and}
  \bibinfo{person}{Nikola~K Kasabov}.} \bibinfo{year}{2022}\natexlab{}.
\newblock \showarticletitle{Learning a coordinated network for
  detail-refinement multi-exposure image fusion}.
\newblock \bibinfo{journal}{\emph{IEEE Transactions on Circuits and Systems for
  Video Technology}} (\bibinfo{year}{2022}).
\newblock


\bibitem[\protect\citeauthoryear{Li, Liu, Zhou, Zhang, and Kasabov}{Li
  et~al\mbox{.}}{2023a}]%
        {li2023gesenet}
\bibfield{author}{\bibinfo{person}{Jiawei Li}, \bibinfo{person}{Jinyuan Liu},
  \bibinfo{person}{Shihua Zhou}, \bibinfo{person}{Qiang Zhang}, {and}
  \bibinfo{person}{Nikola~K Kasabov}.} \bibinfo{year}{2023}\natexlab{a}.
\newblock \showarticletitle{GeSeNet: A General Semantic-Guided Network With
  Couple Mask Ensemble for Medical Image Fusion}.
\newblock \bibinfo{journal}{\emph{IEEE Transactions on Neural Networks and
  Learning Systems}} (\bibinfo{year}{2023}).
\newblock


\bibitem[\protect\citeauthoryear{Li, Liu, Zhou, Zhang, and Kasabov}{Li
  et~al\mbox{.}}{2023b}]%
        {li2023infrared}
\bibfield{author}{\bibinfo{person}{Jiawei Li}, \bibinfo{person}{Jinyuan Liu},
  \bibinfo{person}{Shihua Zhou}, \bibinfo{person}{Qiang Zhang}, {and}
  \bibinfo{person}{Nikola~K Kasabov}.} \bibinfo{year}{2023}\natexlab{b}.
\newblock \showarticletitle{Infrared and visible image fusion based on residual
  dense network and gradient loss}.
\newblock \bibinfo{journal}{\emph{Infrared Physics \& Technology}}
  \bibinfo{volume}{128} (\bibinfo{year}{2023}), \bibinfo{pages}{104486}.
\newblock


\bibitem[\protect\citeauthoryear{Li, Zhang, Shen, Liu, Liu, and Li}{Li
  et~al\mbox{.}}{2021b}]%
        {li2021image}
\bibfield{author}{\bibinfo{person}{Tengpeng Li}, \bibinfo{person}{Kaihua
  Zhang}, \bibinfo{person}{Shiwen Shen}, \bibinfo{person}{Bo Liu},
  \bibinfo{person}{Qingshan Liu}, {and} \bibinfo{person}{Zhu Li}.}
  \bibinfo{year}{2021}\natexlab{b}.
\newblock \showarticletitle{Image co-saliency detection and instance
  co-segmentation using attention graph clustering based graph convolutional
  network}.
\newblock \bibinfo{journal}{\emph{IEEE Transactions on Multimedia}}
  \bibinfo{volume}{24} (\bibinfo{year}{2021}), \bibinfo{pages}{492--505}.
\newblock


\bibitem[\protect\citeauthoryear{Li, Fu, and Zha}{Li et~al\mbox{.}}{2021a}]%
        {li2021cross}
\bibfield{author}{\bibinfo{person}{Yao Li}, \bibinfo{person}{Xueyang Fu}, {and}
  \bibinfo{person}{Zheng-Jun Zha}.} \bibinfo{year}{2021}\natexlab{a}.
\newblock \showarticletitle{Cross-patch graph convolutional network for image
  denoising}. In \bibinfo{booktitle}{\emph{Proceedings of the IEEE/CVF
  International Conference on Computer Vision}}. \bibinfo{pages}{4651--4660}.
\newblock


\bibitem[\protect\citeauthoryear{Liang, Jiang, Liu, and Ma}{Liang
  et~al\mbox{.}}{2022}]%
        {liang2022fusion}
\bibfield{author}{\bibinfo{person}{Pengwei Liang}, \bibinfo{person}{Junjun
  Jiang}, \bibinfo{person}{Xianming Liu}, {and} \bibinfo{person}{Jiayi Ma}.}
  \bibinfo{year}{2022}\natexlab{}.
\newblock \showarticletitle{Fusion from Decomposition: A Self-Supervised
  Decomposition Approach for Image Fusion}. In
  \bibinfo{booktitle}{\emph{Computer Vision--ECCV 2022: 17th European
  Conference, Tel Aviv, Israel, October 23--27, 2022, Proceedings, Part
  XVIII}}. Springer, \bibinfo{pages}{719--735}.
\newblock


\bibitem[\protect\citeauthoryear{Liu, Fan, Huang, Wu, Liu, Zhong, and Luo}{Liu
  et~al\mbox{.}}{2022a}]%
        {liu2022target}
\bibfield{author}{\bibinfo{person}{Jinyuan Liu}, \bibinfo{person}{Xin Fan},
  \bibinfo{person}{Zhanbo Huang}, \bibinfo{person}{Guanyao Wu},
  \bibinfo{person}{Risheng Liu}, \bibinfo{person}{Wei Zhong}, {and}
  \bibinfo{person}{Zhongxuan Luo}.} \bibinfo{year}{2022}\natexlab{a}.
\newblock \showarticletitle{Target-aware dual adversarial learning and a
  multi-scenario multi-modality benchmark to fuse infrared and visible for
  object detection}. In \bibinfo{booktitle}{\emph{Proceedings of the IEEE/CVF
  Conference on Computer Vision and Pattern Recognition}}.
  \bibinfo{pages}{5802--5811}.
\newblock


\bibitem[\protect\citeauthoryear{Liu, Fan, Jiang, Liu, and Luo}{Liu
  et~al\mbox{.}}{2021a}]%
        {liu2021learning}
\bibfield{author}{\bibinfo{person}{Jinyuan Liu}, \bibinfo{person}{Xin Fan},
  \bibinfo{person}{Ji Jiang}, \bibinfo{person}{Risheng Liu}, {and}
  \bibinfo{person}{Zhongxuan Luo}.} \bibinfo{year}{2021}\natexlab{a}.
\newblock \showarticletitle{Learning a deep multi-scale feature ensemble and an
  edge-attention guidance for image fusion}.
\newblock \bibinfo{journal}{\emph{IEEE Transactions on Circuits and Systems for
  Video Technology}} \bibinfo{volume}{32}, \bibinfo{number}{1}
  (\bibinfo{year}{2021}), \bibinfo{pages}{105--119}.
\newblock


\bibitem[\protect\citeauthoryear{Liu, Wu, Luan, Jiang, Liu, and Fan}{Liu
  et~al\mbox{.}}{2023}]%
        {liu2023holoco}
\bibfield{author}{\bibinfo{person}{Jinyuan Liu}, \bibinfo{person}{Guanyao Wu},
  \bibinfo{person}{Junsheng Luan}, \bibinfo{person}{Zhiying Jiang},
  \bibinfo{person}{Risheng Liu}, {and} \bibinfo{person}{Xin Fan}.}
  \bibinfo{year}{2023}\natexlab{}.
\newblock \showarticletitle{HoLoCo: Holistic and local contrastive learning
  network for multi-exposure image fusion}.
\newblock \bibinfo{journal}{\emph{Information Fusion}}  \bibinfo{volume}{95}
  (\bibinfo{year}{2023}), \bibinfo{pages}{237--249}.
\newblock


\bibitem[\protect\citeauthoryear{Liu, Wu, Huang, Liu, and Fan}{Liu
  et~al\mbox{.}}{2021b}]%
        {liu2021smoa}
\bibfield{author}{\bibinfo{person}{Jinyuan Liu}, \bibinfo{person}{Yuhui Wu},
  \bibinfo{person}{Zhanbo Huang}, \bibinfo{person}{Risheng Liu}, {and}
  \bibinfo{person}{Xin Fan}.} \bibinfo{year}{2021}\natexlab{b}.
\newblock \showarticletitle{Smoa: Searching a modality-oriented architecture
  for infrared and visible image fusion}.
\newblock \bibinfo{journal}{\emph{IEEE Signal Processing Letters}}
  \bibinfo{volume}{28} (\bibinfo{year}{2021}), \bibinfo{pages}{1818--1822}.
\newblock


\bibitem[\protect\citeauthoryear{Liu, Wu, Wu, Liu, and Fan}{Liu
  et~al\mbox{.}}{2022b}]%
        {liu2022learn}
\bibfield{author}{\bibinfo{person}{Jinyuan Liu}, \bibinfo{person}{Yuhui Wu},
  \bibinfo{person}{Guanyao Wu}, \bibinfo{person}{Risheng Liu}, {and}
  \bibinfo{person}{Xin Fan}.} \bibinfo{year}{2022}\natexlab{b}.
\newblock \showarticletitle{Learn to Search a Lightweight Architecture for
  Target-Aware Infrared and Visible Image Fusion}.
\newblock \bibinfo{journal}{\emph{IEEE Signal Processing Letters}}
  \bibinfo{volume}{29} (\bibinfo{year}{2022}), \bibinfo{pages}{1614--1618}.
\newblock


\bibitem[\protect\citeauthoryear{Liu, Liu, Jiang, Fan, and Luo}{Liu
  et~al\mbox{.}}{2020}]%
        {liu2020bilevel}
\bibfield{author}{\bibinfo{person}{Risheng Liu}, \bibinfo{person}{Jinyuan Liu},
  \bibinfo{person}{Zhiying Jiang}, \bibinfo{person}{Xin Fan}, {and}
  \bibinfo{person}{Zhongxuan Luo}.} \bibinfo{year}{2020}\natexlab{}.
\newblock \showarticletitle{A bilevel integrated model with data-driven layer
  ensemble for multi-modality image fusion}.
\newblock \bibinfo{journal}{\emph{IEEE Transactions on Image Processing}}
  \bibinfo{volume}{30} (\bibinfo{year}{2020}), \bibinfo{pages}{1261--1274}.
\newblock


\bibitem[\protect\citeauthoryear{Luo, Li, Yang, Jiao, Cheng, and Lyu}{Luo
  et~al\mbox{.}}{2020}]%
        {luo2020cascade}
\bibfield{author}{\bibinfo{person}{Ao Luo}, \bibinfo{person}{Xin Li},
  \bibinfo{person}{Fan Yang}, \bibinfo{person}{Zhicheng Jiao},
  \bibinfo{person}{Hong Cheng}, {and} \bibinfo{person}{Siwei Lyu}.}
  \bibinfo{year}{2020}\natexlab{}.
\newblock \showarticletitle{Cascade graph neural networks for RGB-D salient
  object detection}. In \bibinfo{booktitle}{\emph{Computer Vision--ECCV 2020:
  16th European Conference, Glasgow, UK, August 23--28, 2020, Proceedings, Part
  XII 16}}. Springer, \bibinfo{pages}{346--364}.
\newblock


\bibitem[\protect\citeauthoryear{Ma, Tang, Fan, Huang, Mei, and Ma}{Ma
  et~al\mbox{.}}{2022}]%
        {ma2022swinfusion}
\bibfield{author}{\bibinfo{person}{Jiayi Ma}, \bibinfo{person}{Linfeng Tang},
  \bibinfo{person}{Fan Fan}, \bibinfo{person}{Jun Huang},
  \bibinfo{person}{Xiaoguang Mei}, {and} \bibinfo{person}{Yong Ma}.}
  \bibinfo{year}{2022}\natexlab{}.
\newblock \showarticletitle{SwinFusion: Cross-domain long-range learning for
  general image fusion via swin transformer}.
\newblock \bibinfo{journal}{\emph{IEEE/CAA Journal of Automatica Sinica}}
  \bibinfo{volume}{9}, \bibinfo{number}{7} (\bibinfo{year}{2022}),
  \bibinfo{pages}{1200--1217}.
\newblock


\bibitem[\protect\citeauthoryear{Ma, Xu, Jiang, Mei, and Zhang}{Ma
  et~al\mbox{.}}{2020}]%
        {ma2020ddcgan}
\bibfield{author}{\bibinfo{person}{Jiayi Ma}, \bibinfo{person}{Han Xu},
  \bibinfo{person}{Junjun Jiang}, \bibinfo{person}{Xiaoguang Mei}, {and}
  \bibinfo{person}{Xiao-Ping Zhang}.} \bibinfo{year}{2020}\natexlab{}.
\newblock \showarticletitle{DDcGAN: A dual-discriminator conditional generative
  adversarial network for multi-resolution image fusion}.
\newblock \bibinfo{journal}{\emph{IEEE Transactions on Image Processing}}
  \bibinfo{volume}{29} (\bibinfo{year}{2020}), \bibinfo{pages}{4980--4995}.
\newblock


\bibitem[\protect\citeauthoryear{Paramanandham and Rajendiran}{Paramanandham
  and Rajendiran}{2018}]%
        {paramanandham2018infrared}
\bibfield{author}{\bibinfo{person}{Nirmala Paramanandham} {and}
  \bibinfo{person}{Kishore Rajendiran}.} \bibinfo{year}{2018}\natexlab{}.
\newblock \showarticletitle{Infrared and visible image fusion using discrete
  cosine transform and swarm intelligence for surveillance applications}.
\newblock \bibinfo{journal}{\emph{Infrared Physics \& Technology}}
  \bibinfo{volume}{88} (\bibinfo{year}{2018}), \bibinfo{pages}{13--22}.
\newblock


\bibitem[\protect\citeauthoryear{Redmon, Divvala, Girshick, and Farhadi}{Redmon
  et~al\mbox{.}}{2016}]%
        {redmon2016you}
\bibfield{author}{\bibinfo{person}{Joseph Redmon}, \bibinfo{person}{Santosh
  Divvala}, \bibinfo{person}{Ross Girshick}, {and} \bibinfo{person}{Ali
  Farhadi}.} \bibinfo{year}{2016}\natexlab{}.
\newblock \showarticletitle{You only look once: Unified, real-time object
  detection}. In \bibinfo{booktitle}{\emph{Proceedings of the IEEE conference
  on computer vision and pattern recognition}}. \bibinfo{pages}{779--788}.
\newblock


\bibitem[\protect\citeauthoryear{Shah, Merchant, and Desai}{Shah
  et~al\mbox{.}}{2013}]%
        {shah2013multifocus}
\bibfield{author}{\bibinfo{person}{Parul Shah}, \bibinfo{person}{Shabbir~N
  Merchant}, {and} \bibinfo{person}{Uday~B Desai}.}
  \bibinfo{year}{2013}\natexlab{}.
\newblock \showarticletitle{Multifocus and multispectral image fusion based on
  pixel significance using multiresolution decomposition}.
\newblock \bibinfo{journal}{\emph{Signal, Image and Video Processing}}
  \bibinfo{volume}{7} (\bibinfo{year}{2013}), \bibinfo{pages}{95--109}.
\newblock


\bibitem[\protect\citeauthoryear{Song, Huang, Gong, and Yan}{Song
  et~al\mbox{.}}{2022}]%
        {song2022multiple}
\bibfield{author}{\bibinfo{person}{Kechen Song}, \bibinfo{person}{Liming
  Huang}, \bibinfo{person}{Aojun Gong}, {and} \bibinfo{person}{Yunhui Yan}.}
  \bibinfo{year}{2022}\natexlab{}.
\newblock \showarticletitle{Multiple graph affinity interactive network and a
  variable illumination dataset for RGBT image salient object detection}.
\newblock \bibinfo{journal}{\emph{IEEE Transactions on Circuits and Systems for
  Video Technology}} (\bibinfo{year}{2022}).
\newblock


\bibitem[\protect\citeauthoryear{Sun, Cao, Zhu, and Hu}{Sun
  et~al\mbox{.}}{2022a}]%
        {sun2022detfusion}
\bibfield{author}{\bibinfo{person}{Yiming Sun}, \bibinfo{person}{Bing Cao},
  \bibinfo{person}{Pengfei Zhu}, {and} \bibinfo{person}{Qinghua Hu}.}
  \bibinfo{year}{2022}\natexlab{a}.
\newblock \showarticletitle{Detfusion: A detection-driven infrared and visible
  image fusion network}. In \bibinfo{booktitle}{\emph{Proceedings of the 30th
  ACM International Conference on Multimedia}}. \bibinfo{pages}{4003--4011}.
\newblock


\bibitem[\protect\citeauthoryear{Sun, Cao, Zhu, and Hu}{Sun
  et~al\mbox{.}}{2022b}]%
        {sun2022drone}
\bibfield{author}{\bibinfo{person}{Yiming Sun}, \bibinfo{person}{Bing Cao},
  \bibinfo{person}{Pengfei Zhu}, {and} \bibinfo{person}{Qinghua Hu}.}
  \bibinfo{year}{2022}\natexlab{b}.
\newblock \showarticletitle{Drone-based RGB-infrared cross-modality vehicle
  detection via uncertainty-aware learning}.
\newblock \bibinfo{journal}{\emph{IEEE Transactions on Circuits and Systems for
  Video Technology}} \bibinfo{volume}{32}, \bibinfo{number}{10}
  (\bibinfo{year}{2022}), \bibinfo{pages}{6700--6713}.
\newblock


\bibitem[\protect\citeauthoryear{Tang, Xiang, Zhang, Gong, and Ma}{Tang
  et~al\mbox{.}}{2023}]%
        {tang2023divfusion}
\bibfield{author}{\bibinfo{person}{Linfeng Tang}, \bibinfo{person}{Xinyu
  Xiang}, \bibinfo{person}{Hao Zhang}, \bibinfo{person}{Meiqi Gong}, {and}
  \bibinfo{person}{Jiayi Ma}.} \bibinfo{year}{2023}\natexlab{}.
\newblock \showarticletitle{DIVFusion: Darkness-free infrared and visible image
  fusion}.
\newblock \bibinfo{journal}{\emph{Information Fusion}}  \bibinfo{volume}{91}
  (\bibinfo{year}{2023}), \bibinfo{pages}{477--493}.
\newblock


\bibitem[\protect\citeauthoryear{Tang, Yuan, and Ma}{Tang
  et~al\mbox{.}}{2022}]%
        {tang2022image}
\bibfield{author}{\bibinfo{person}{Linfeng Tang}, \bibinfo{person}{Jiteng
  Yuan}, {and} \bibinfo{person}{Jiayi Ma}.} \bibinfo{year}{2022}\natexlab{}.
\newblock \showarticletitle{Image fusion in the loop of high-level vision
  tasks: A semantic-aware real-time infrared and visible image fusion network}.
\newblock \bibinfo{journal}{\emph{Information Fusion}}  \bibinfo{volume}{82}
  (\bibinfo{year}{2022}), \bibinfo{pages}{28--42}.
\newblock


\bibitem[\protect\citeauthoryear{Toet}{Toet}{2017}]%
        {toet2017tno}
\bibfield{author}{\bibinfo{person}{Alexander Toet}.}
  \bibinfo{year}{2017}\natexlab{}.
\newblock \showarticletitle{The TNO multiband image data collection}.
\newblock \bibinfo{journal}{\emph{Data in brief}}  \bibinfo{volume}{15}
  (\bibinfo{year}{2017}), \bibinfo{pages}{249--251}.
\newblock


\bibitem[\protect\citeauthoryear{Wang, Liu, Fan, and Liu}{Wang
  et~al\mbox{.}}{2022}]%
        {wang2022unsupervised}
\bibfield{author}{\bibinfo{person}{Di Wang}, \bibinfo{person}{Jinyuan Liu},
  \bibinfo{person}{Xin Fan}, {and} \bibinfo{person}{Risheng Liu}.}
  \bibinfo{year}{2022}\natexlab{}.
\newblock \showarticletitle{Unsupervised misaligned infrared and visible image
  fusion via cross-modality image generation and registration}.
\newblock \bibinfo{journal}{\emph{arXiv preprint arXiv:2205.11876}}
  (\bibinfo{year}{2022}).
\newblock


\bibitem[\protect\citeauthoryear{Wang, Bovik, Sheikh, and Simoncelli}{Wang
  et~al\mbox{.}}{2004}]%
        {wang2004image}
\bibfield{author}{\bibinfo{person}{Zhou Wang}, \bibinfo{person}{Alan~C Bovik},
  \bibinfo{person}{Hamid~R Sheikh}, {and} \bibinfo{person}{Eero~P Simoncelli}.}
  \bibinfo{year}{2004}\natexlab{}.
\newblock \showarticletitle{Image quality assessment: from error visibility to
  structural similarity}.
\newblock \bibinfo{journal}{\emph{IEEE transactions on image processing}}
  \bibinfo{volume}{13}, \bibinfo{number}{4} (\bibinfo{year}{2004}),
  \bibinfo{pages}{600--612}.
\newblock


\bibitem[\protect\citeauthoryear{Xie, Liu, Xiong, and Shao}{Xie
  et~al\mbox{.}}{2021}]%
        {xie2021scale}
\bibfield{author}{\bibinfo{person}{Guo-Sen Xie}, \bibinfo{person}{Jie Liu},
  \bibinfo{person}{Huan Xiong}, {and} \bibinfo{person}{Ling Shao}.}
  \bibinfo{year}{2021}\natexlab{}.
\newblock \showarticletitle{Scale-aware graph neural network for few-shot
  semantic segmentation}. In \bibinfo{booktitle}{\emph{Proceedings of the
  IEEE/CVF conference on computer vision and pattern recognition}}.
  \bibinfo{pages}{5475--5484}.
\newblock


\bibitem[\protect\citeauthoryear{Xu, Gong, Tian, Huang, and Ma}{Xu
  et~al\mbox{.}}{2022}]%
        {xu2022cufd}
\bibfield{author}{\bibinfo{person}{Han Xu}, \bibinfo{person}{Meiqi Gong},
  \bibinfo{person}{Xin Tian}, \bibinfo{person}{Jun Huang}, {and}
  \bibinfo{person}{Jiayi Ma}.} \bibinfo{year}{2022}\natexlab{}.
\newblock \showarticletitle{CUFD: An encoder--decoder network for visible and
  infrared image fusion based on common and unique feature decomposition}.
\newblock \bibinfo{journal}{\emph{Computer Vision and Image Understanding}}
  \bibinfo{volume}{218} (\bibinfo{year}{2022}), \bibinfo{pages}{103407}.
\newblock


\bibitem[\protect\citeauthoryear{Xu, Ma, Jiang, Guo, and Ling}{Xu
  et~al\mbox{.}}{2020}]%
        {xu2020u2fusion}
\bibfield{author}{\bibinfo{person}{Han Xu}, \bibinfo{person}{Jiayi Ma},
  \bibinfo{person}{Junjun Jiang}, \bibinfo{person}{Xiaojie Guo}, {and}
  \bibinfo{person}{Haibin Ling}.} \bibinfo{year}{2020}\natexlab{}.
\newblock \showarticletitle{U2Fusion: A unified unsupervised image fusion
  network}.
\newblock \bibinfo{journal}{\emph{IEEE Transactions on Pattern Analysis and
  Machine Intelligence}} \bibinfo{volume}{44}, \bibinfo{number}{1}
  (\bibinfo{year}{2020}), \bibinfo{pages}{502--518}.
\newblock


\bibitem[\protect\citeauthoryear{Xydeas, Petrovic, et~al\mbox{.}}{Xydeas
  et~al\mbox{.}}{2000}]%
        {xydeas2000objective}
\bibfield{author}{\bibinfo{person}{Costas~S Xydeas}, \bibinfo{person}{Vladimir
  Petrovic}, {et~al\mbox{.}}} \bibinfo{year}{2000}\natexlab{}.
\newblock \showarticletitle{Objective image fusion performance measure}.
\newblock \bibinfo{journal}{\emph{Electronics letters}} \bibinfo{volume}{36},
  \bibinfo{number}{4} (\bibinfo{year}{2000}), \bibinfo{pages}{308--309}.
\newblock


\bibitem[\protect\citeauthoryear{Zhang and Ma}{Zhang and Ma}{2021}]%
        {zhang2021sdnet}
\bibfield{author}{\bibinfo{person}{Hao Zhang} {and} \bibinfo{person}{Jiayi
  Ma}.} \bibinfo{year}{2021}\natexlab{}.
\newblock \showarticletitle{SDNet: A versatile squeeze-and-decomposition
  network for real-time image fusion}.
\newblock \bibinfo{journal}{\emph{International Journal of Computer Vision}}
  \bibinfo{volume}{129} (\bibinfo{year}{2021}), \bibinfo{pages}{2761--2785}.
\newblock


\bibitem[\protect\citeauthoryear{Zhang, Xu, Tian, Jiang, and Ma}{Zhang
  et~al\mbox{.}}{2021}]%
        {zhang2021image}
\bibfield{author}{\bibinfo{person}{Hao Zhang}, \bibinfo{person}{Han Xu},
  \bibinfo{person}{Xin Tian}, \bibinfo{person}{Junjun Jiang}, {and}
  \bibinfo{person}{Jiayi Ma}.} \bibinfo{year}{2021}\natexlab{}.
\newblock \showarticletitle{Image fusion meets deep learning: A survey and
  perspective}.
\newblock \bibinfo{journal}{\emph{Information Fusion}}  \bibinfo{volume}{76}
  (\bibinfo{year}{2021}), \bibinfo{pages}{323--336}.
\newblock


\bibitem[\protect\citeauthoryear{Zhang, Jiao, Ma, Liu, Liu, Li, Chen, and
  Yang}{Zhang et~al\mbox{.}}{2023}]%
        {zhang2023transformer}
\bibfield{author}{\bibinfo{person}{Jun Zhang}, \bibinfo{person}{Licheng Jiao},
  \bibinfo{person}{Wenping Ma}, \bibinfo{person}{Fang Liu}, \bibinfo{person}{Xu
  Liu}, \bibinfo{person}{Lingling Li}, \bibinfo{person}{Puhua Chen}, {and}
  \bibinfo{person}{Shuyuan Yang}.} \bibinfo{year}{2023}\natexlab{}.
\newblock \showarticletitle{Transformer based Conditional GAN for Multimodal
  Image Fusion}.
\newblock \bibinfo{journal}{\emph{IEEE Transactions on Multimedia}}
  (\bibinfo{year}{2023}).
\newblock


\bibitem[\protect\citeauthoryear{Zhao, Shi, Qi, Wang, and Jia}{Zhao
  et~al\mbox{.}}{2017}]%
        {zhao2017pyramid}
\bibfield{author}{\bibinfo{person}{Hengshuang Zhao}, \bibinfo{person}{Jianping
  Shi}, \bibinfo{person}{Xiaojuan Qi}, \bibinfo{person}{Xiaogang Wang}, {and}
  \bibinfo{person}{Jiaya Jia}.} \bibinfo{year}{2017}\natexlab{}.
\newblock \showarticletitle{Pyramid scene parsing network}. In
  \bibinfo{booktitle}{\emph{Proceedings of the IEEE conference on computer
  vision and pattern recognition}}. \bibinfo{pages}{2881--2890}.
\newblock


\bibitem[\protect\citeauthoryear{Zhao and Bai}{Zhao and Bai}{2022}]%
        {zhao2022cddfuse}
\bibfield{author}{\bibinfo{person}{Zixiang Zhao} {and} \bibinfo{person}{Haowen
  et~al. Bai}.} \bibinfo{year}{2022}\natexlab{}.
\newblock \showarticletitle{CDDFuse: Correlation-Driven Dual-Branch Feature
  Decomposition for Multi-Modality Image Fusion}.
\newblock \bibinfo{journal}{\emph{arXiv preprint arXiv:2211.14461}}
  (\bibinfo{year}{2022}).
\newblock


\bibitem[\protect\citeauthoryear{Zhao and Xu}{Zhao and Xu}{2020}]%
        {zhao2020didfuse}
\bibfield{author}{\bibinfo{person}{Zixiang Zhao} {and} \bibinfo{person}{Shuang
  et~al. Xu}.} \bibinfo{year}{2020}\natexlab{}.
\newblock \showarticletitle{DIDFuse: Deep image decomposition for infrared and
  visible image fusion}.
\newblock \bibinfo{journal}{\emph{arXiv preprint arXiv:2003.09210}}
  (\bibinfo{year}{2020}).
\newblock


\end{thebibliography}

\end{document}